\documentclass[preprint,12pt,authoryear]{elsarticle}

\usepackage{amsmath,amsfonts}
\usepackage{amssymb}
\usepackage{algorithmic}
\usepackage{algorithm}
\usepackage{array}
\usepackage{multirow}%
\usepackage{multicol}
\usepackage{adjustbox}
\usepackage[caption=false,font=normalsize,labelfont=sf,textfont=sf]{subfig}
\usepackage{textcomp}
\usepackage{stfloats}
\usepackage{url}
\usepackage{verbatim}
\usepackage{graphicx}
\usepackage{cite}
\usepackage{hyperref}
\journal{Neural Networks}

\begin{document}

\begin{frontmatter}



\title{A Novel Energy-Efficient Spike Transformer Network for Depth
Estimation from Event Cameras via Cross-modality Knowledge Distillation} 

\author[1]{Xin Zhang}
\author[1]{Liangxiu Han\corref{cor1}}
\ead{l.han@mmu.ac.uk}
\cortext[cor1]{Corresponding author}
\author[1]{Tam Sobeih}
\author[2]{Lianghao Han}
\author[1]{Darren Dancey}

\address[1]{Department of Computing, and Mathematics, Manchester Metropolitan University, Manchester M15 6BH, UK}
\address[2]{Department of Computer Science,  Brunel University, Uxbridge UB8 3PH, UK}


\begin{abstract}
Depth estimation is a critical task in computer vision, with applications in autonomous navigation, robotics, and augmented reality. Event cameras, which encode temporal changes in light intensity as asynchronous binary spikes, offer unique advantages such as low latency, high dynamic range, and energy efficiency. However, their unconventional spiking output and the scarcity of labelled datasets pose significant challenges to traditional image-based depth estimation methods. To address these challenges, we propose a novel energy-efficient Spike-Driven Transformer Network (SDT) for depth estimation, leveraging the unique properties of spiking data. The proposed SDT introduces three key innovations: (1) a purely spike-driven transformer architecture that incorporates spike-based attention and residual mechanisms, enabling precise depth estimation with minimal energy consumption; (2) a fusion depth estimation head that combines multi-stage features for fine-grained depth prediction while ensuring computational efficiency; and (3) a cross-modality knowledge distillation framework that utilises a pre-trained vision foundation model (DINOv2) to enhance the training of the spiking network despite limited data availability.
Experimental evaluations on synthetic and real-world event datasets demonstrate the superiority of our approach, with substantial improvements in Absolute Relative Error (49\% reduction) and Square Relative Error (39.77\% reduction) compared to existing models. The SDT also achieves a 70.2\% reduction in energy consumption (12.43 mJ vs. 41.77 mJ per inference) and reduces model parameters by 42.4\% (20.55M vs. 35.68M), making it highly suitable for resource-constrained environments. This work represents the first exploration of transformer-based spiking neural networks for depth estimation, providing a significant step forward in energy-efficient neuromorphic computing for real-world vision applications.

\end{abstract}




\end{frontmatter}




\section{Introduction}

Event-based cameras are bio-inspired sensors that capture visual information asynchronously, reporting brightness changes in real-time \citep{Gallego2022Event-Based,Tayarani-Najaran2021Event-Based}. Due to hardware design, they offer many advantages over traditional cameras, including low latency between trigger events \citep{Chen2019Live}, low power consumption \citep{Posch2011QVGA}, and high dynamic range \citep{Furmonas2022Analytical}. These event-based sensors are already being used in a variety of fields such as 3D scanning \citep{Huang2022Real-time}, robotic vision \citep{Cao2021Fusion-based}, and the automotive industry\citep{Chen2020Event-Based}. In practice, event-based sensors capture unique spiking data, encoding information about changes in light intensity in a scene. The noise in the data is extremely high and there is a lack of generalized processing algorithms to provide capabilities comparable to traditional vision algorithms for conventional digital camera data.

Spiking Neural Networks (SNNs) are bioinspired neural network models that mimic the behavior of biological neural networks. SNNs use discrete functions called spikes to represent and process information, as opposed to continuous values used in traditional vision algorithms \citep{Rafi2021brief}. Therefore, it is naturally an ideal paradigm for processing the output of event cameras \citep{Lee2020Spike-FlowNet,Auge2021Survey,Cordone2021Learning}.

Depth estimation is a challenging task in computer vision, used widely in autonomous driving, automated robotics, agricultural growth monitoring, and forest carbon emission monitoring. Current state-of-the-art depth prediction efforts have mainly focused on combining standard frame-based cameras with artificial neural networks (ANN) \citep{Zhao2020Monocular,Ming2021Deep,Laga2022Survey}. Event-based cameras and associated processing algorithms for depth-sensing applications are still in their infancy \citep{Furmonas2022Analytical}. Two major challenges include the lack of a SNN backbone for feature extraction and poor SNN model performance.

\textbf{The lack of SNN backbone designed for spike data depth estimation}. The event-based camera generates continuous spike streams in a binary irregular data structure that possesses ultrahigh temporal features. SNNs are applicable to event camera datasets and able to improve the depth estimation performance by exploiting advanced architecture of ANN, such as ResNet like SNNs and Spiking Recurrent Neural network \citep{Fang2021Deep,Hu2023Spiking,Yin2020Effective,Diehl2016Conversion}. Vision transformer  \citep{Dosovitskiy2020Image,Han2023Survey}(ViT), is currently the most popular ANN structure and is based on a self-attention mechanism to capture long-distance dependencies, especially spatial-temporal features in images/videos. It improves the performance of AI in many computer vision tasks such as image classification/ segmentation \citep{Strudel2021Segmenter:,Arnab2021ViViT:,Liu2021Swin}, object detection \citep{Sun2021Rethinking} and depth estimation \citep{Yang2022Depth,Zhao2022MonoViT:}. Transformer-based SNNs are a new form of SNN combining transformer architectures with spiking neurons, offering great potential to break the performance bottleneck on spike stream data. In \citet{Zhang2022Spike}, the authors used the original ViT structure as a backbone to extract features from both spatial and temporal domains in spike data. The result demonstrated the suitability of the transformer for extracting spatio-temporal features. However, the original transformer structure has a large number of multiplication operations and excessive computational energy consumption compared to SNNs. In  \citet{Zhou2022Spikformer:} and \citet{Zhou2023Spikingformer:}, the author proposed a pure spike driven self-attention and residual connection to avoid non-spike computations. This was a major step forward in the potential use of transformer for depth estimation from spike data.

\textbf{SNN model performance}. One of the biggest challenges with SNNs currently is their inability to achieve equivalent performance on spiking data to ANNs on non-spiking data. Gradient-based backpropagation is a powerful algorithm for training ANNs, but as spiking data is non-differentiable it cannot be used directly with SNNs \citep{He2023Improving}. Converting ANN to SNN is a solution but it may introduce errors of uncertainty or lose the temporal information of spikes \citep{Diehl2016Conversion}. Meanwhile, the number of event-based datasets are small compared to the static images used in traditional ANN training, making SNNs prone to overfitting and limiting their generalization ability \citep{He2023Improving}. Knowledge distillation is a technique in deep learning to transfer knowledge from the teacher model to the student model. It allows training of a lightweight model (student model) to be as accurate as a larger model (teacher model). Currently, there are already some ANN models trained with massive data that can achieve zero-shot for depth estimation \citep{Oquab2023DINOv2:,Ranftl2020Towards,Ranftl2021Vision}. Logically, the accuracy of these models has the potential to be transferred to the SNN model during training.

In this work, we propose a novel energy-efficient spike transformer  network for depth estimation, leveraging cross-modality knowledge distillation to combine the biological efficiency of SNNs with the advanced feature extraction capabilities of a visual foundation model (DINOv2). To the best of our knowledge, this is the first exploration of a transformer-based SNN for depth estimation, marking a significant advancement in the field. The proposed framework comprises three key components, each contributing uniquely to its overall effectiveness:

\begin{enumerate}

\item [1)] We introduce a novel energy-efficient spike-driven transformer that eliminates conventional floating-point operations through carefully designed spike-based attention and residual mechanisms. This network incorporates two essential components: a spiking patch embedding module that converts raw event data into spike-based tokens while preserving temporal-spatial information, and spiking transformer blocks that integrate Spiking Self-Attention (SSA) and Spiking MLP for efficient feature processing. This design significantly reduces energy consumption while ensuring robust performance.

\item [2)] We develop a fusion depth estimation head that combines features from multiple transformer stages to enable fine-grained depth prediction while maintaining efficient computation. Unlike conventional CNN-based methods that rely on downsampling, our fusion head preserves critical spatial information through a multi-scale feature integration approach, proving particularly effective in challenging scenarios such as low-light conditions and distant object detection.

\item [3)] We propose a single-stage cross-modality knowledge distillation framework that leverages a large vision foundation model (DINOv2) to enhance SNN training with limited data. By utilis ing domain loss and semantic loss, our framework effectively transfers knowledge from both final and intermediate layers of DINOv2 to the spike-driven transformer. 


\end{enumerate}

\section{Related works}

This section presents a literature review of existing research in monocular depth estimation, SNNs, and knowledge distillation, highlighting the key challenges that motivate our work.

\subsection{Image-based and Event-based Monocular Depth Estimations}

Depth estimation from images aims to measure the distance of each pixel relative to the camera. Monocular depth estimation is a challenging but promising technology. It has the advantage of only requiring one image unlike traditional depth estimation, which makes it more practical for applications where it is not possible to take a pair of images, such as mobile devices. Depending on the type of data used, we can divide  monocular depth estimation into Image-based and Event-based methods \citep{Ming2021Deep}. Image-based monocular depth estimation is more common as it estimates depth using the information in RGB images, which are easy to collect and process. The event-based method uses spike data generated by an event camera, a new type of sensor that outputs brightness changes in the form of an asynchronous ”event” stream, rather than static images. This makes them well-suited for depth estimation in challenging conditions, such as low light and fast motion \citep{Cordone2021Learning}, but the data is harder to collect and process.

The latest developments in deep learning have made it possible to develop monocular depth estimation models that can achieve satisfactory accuracy and robustness \citep{Ming2021Deep,Khan2020Deep,Li2020Deep}. Similar to other deep learning models, these models typically consist of a generalized encoder that extracts abstract features from context information and a decoder that recovers depth information from the features. For RGB images, in  \citet{Laina2016Deeper}, the author used ResNet-50 as an encoder and novel up-sampling blocks as a decoder to estimate depth from a single RGB image. In \citet{Laina2016Deeper}, the authors utilised ViT instead of convolutional networks as the backbone for a depth estimation task. Experiments have found that transformer is able to provide finer and more globally consistent predictions than traditional convolutional networks. For event data, the research is still in its infancy. The authors \citep{Cordone2021Learning} presented a new deep learning model called E2Depth that can estimate depth from event cameras with high accuracy. A fully convolutional neural network based on the U-Net architecture  \citep{Ronneberger2015U-Net:}  was used in this work. In \citet{Nam2022Stereo}, a multiscale encoder was used to extract features from mixed-density event stacking and an upscaling decoder was used to predict the depth. The transformer structure has also been used in event-based monocular depth estimation. In \citet{Liu2022Event-based}, EReFormer was proposed to estimate depth from event cameras with superior accuracy based on transformer.

However, these models predominantly utilise traditional deep learning frameworks, overlooking the unique potential of event-based data. Existing research identifies two key challenges that remain unaddressed. 
The first challenge lies in the unique characteristics of event camera data. These cameras produce continuous streams of sparse, asynchronous data, which traditional frame-based methods struggle to process efficiently. They also provide very high temporal resolution, capturing subtle movements and rapid changes in brightness. This requires algorithms that can process data in real-time and maintain temporal accuracy \citep{Gallego2022Event-Based,Tayarani-Najaran2021Event-Based}. SNNs are biologically inspired networks that process information using spikes, similar to the human brain, and are well-suited to handle the event-driven nature of these cameras \citep{tavanaei2019deep}. In this paper, we review the current SNN methods in the literature and propose a novel spike transformer network for depth estimation from event camera data.
The second challenge is the scarcity of spike training data. High-quality, labelled datasets tailored for SNNs, particularly for tasks like depth estimation, remain limited. The acquisition and labelling of event-based data are both complex and resource-intensive, further constraining the availability of training resources. To address this limitation, knowledge distillation offers a promising solution. This involves transferring knowledge from a well-trained artificial neural network (ANN). The ANN acts as a ”teacher”, guiding the SNN, or ”student”, to learn effectively with limited event-based data.

\subsection{Spiking Neural Networks (SNNs)}

Unlike traditional deep learning models that convey information using continuous decimal values, SNNs use discrete spike sequences to calculate and transmit information. Spiking neurons receive continuous values and convert them into spike sequences. A number of different spiking neuron models have been proposed. The Hodgkin-Huxley model is one of the first models that describes the behaviour of biological neurons \citep{Hodgkin1952quantitative}, and is fundamental to explain how spikes flow in neurons, but the model is too complex to implement in silicon. The Lzhikevich model \citep{Izhikevich2003Simple}, which simplifies the Hodgkin-Huxley model, is a two-dimensional model that describes the dynamics of the membrane potential of a neuron. The leaky integrate-and-fire (LIF) neuron is another simple neuron model that is widely used in neuroscience and SNNs. It is simpler than Lzhikevich model but captures the essential features on how neurons work. It can be used to build SNNs and implemented in very-large-scale integrations (VLSI)\citep{hsieh2012vlsi}. The membrane potential of the LIF neuron is governed by the following equation:

\begin{equation}
 d v / d t=I-v / \tau 
\end{equation}

where v is the membrane potential, t is time, $\tau$ is a time constant, and I is the current. The current, I, can be either excitatory or inhibitory. Excitatory currents make the membrane potential more positive, while inhibitory currents make it more negative. When the membrane potential reaches the threshold, the neuron fires an action potential. In this work, LIF is used to build the proposed model.

Similar to ANNs, as the depth of SNNs increases, their performance significantly improves \citep{Fang2021Deep,Hu2023Spiking,Zheng2020Going}. Currently, most SNNs have borrowed structures from ANNs, which can be categorised into two main groups: CNN-based SNNs and transformer-based SNNs.
ResNet, as the most successful CNN model has been extensively studied to extend the depth of SNNs \citep{Fang2021Deep,Hu2023Spiking}. SEW ResNet \citep{Fang2021Deep} overcomes the vanishing/exploding gradient problem in SNNs by using a technique called spike-timing dependent plasticity (STDP). It has been shown effective in a variety of tasks, including image classification and object detection. However, Convolutional networks possess translation invariant and locally dependent, but their calculation has a fixed receptive field, limiting their ability to capture global dependence. In contrast, ViT is based on self-attention mechanisms that can capture long-distance dependencies. They are based on the Transformer architecture, which was originally developed for natural language processing tasks.
Transformer-based SNNs represent a novel form of SNNs that combines the transformer architecture with SNNs, providing great potential to break through the performance bottleneck of SNNs.
\citet{Yao2023Spike-driven}, and \citet{Zhou2022Spikformer:}, proposed two different Spike-Driven Self-Attention models. To avoid multiplication, they utilised only mask and addition operations, which are efficient and have low computational energy consumption. \citet{Zhou2023Spikingformer:}, proposed Spikingformer, modifying the residual connection to be purely event-driven, making it energy efficient while improving performance.

Currently, ViTs have been shown to the achieve state-of-the-art results in a variety of vision tasks, including image classification, object detection, segmentation, and depth estimation. However, transformer based SNNs are still in their infancy regarding depth estimation \citep{Zhang2022Spike}. There are two major issues: 1) the difficulty in training pure transformer based SNN models; 2) the limited availability of paired depth data in event data to support transformer model training. Knowledge distillation provides a manner to train new or small models using well pretrained models. In this work, we propose a knowledge distillation method to bring ANN model knowledge into SNN.

\subsection{Knowledge distillation for SNN}

Knowledge distillation is a type of model compression technique that transfers knowledge from large teacher models to smaller student models. It has gained attention due to its ability to train deep neural networks using limited resources \citep{Gou2021Knowledge}, and has been shown effective in improving the performance of SNNs. In \citet{Kushawaha2020Distilling}, the author proposed a knowledge distillation method for transferring knowledge from a large trained SNN to a small one in an image classification task. The results show that using the knowledge of pre-trained large models can significantly improve the performance of small models, thus enhancing the possibility of deploying high-performance models on resource-limited platforms. In study \citep{He2023Improving}, a knowledge distillation method was proposed for SNNs performing image classification. The method was able to improve the accuracy of a student SNN by 2.7\% to 9.8\%. The authors \citep{Qiu2022Self-Architectural}  found that the knowledge distillation paradigm can be effective in reducing the performance gap from the ANN to the SNN. The distillation between ANNs and SNNs using similar structures improved model performance. In \citep{Liu2022Unsupervised}, knowledge distillation was first used in training SNNs for depth estimation. The author proposed a cross-modality domain knowledge transfer method for unsupervised spike depth estimation with open-source RGB data.

While knowledge distillation has shown promise in improving SNN performance \citep{He2023Improving, Qiu2022Self-Architectural}, existing approaches have several limitations: 1) Most methods focus on classification tasks and cannot be directly applied to dense prediction problems like depth estimation; 2) Cross-modality knowledge transfer between conventional images and event data remains largely unexplored; 3) Current approaches typically require training a separate teacher model, increasing computational overhead. Currently, large foundation models have become the new deep learning hotspot \citep{Bommasani2022On}. A large foundation model is trained on a vast quantity of data at scale (often by self-supervised learning or semi-supervised learning) so that the learned features can be used directly for various downstream tasks or knowledge distillation. For example, Dense Prediction Transformers (DPT) \citep{Ranftl2021Vision} are a type of ViT designed for depth prediction tasks, trained on 1.4 million images for monocular depth estimation. DINOv2 \citep{Oquab2023DINOv2:}  used ViT-Giant, a larger version of ViT with 1 billion parameters. It is more powerful than previous ViT models and outperforms previous self-supervised learning methods in a variety of computer vision tasks, especially for depth estimation. In this work, for the first time, we will explore transferring knowledge from a large foundation model (DINOv2) to SNNs for depth estimation.

\section{The Proposed Method}

\begin{figure*}[!htbp]
\centering
\includegraphics[width=0.9\textwidth]{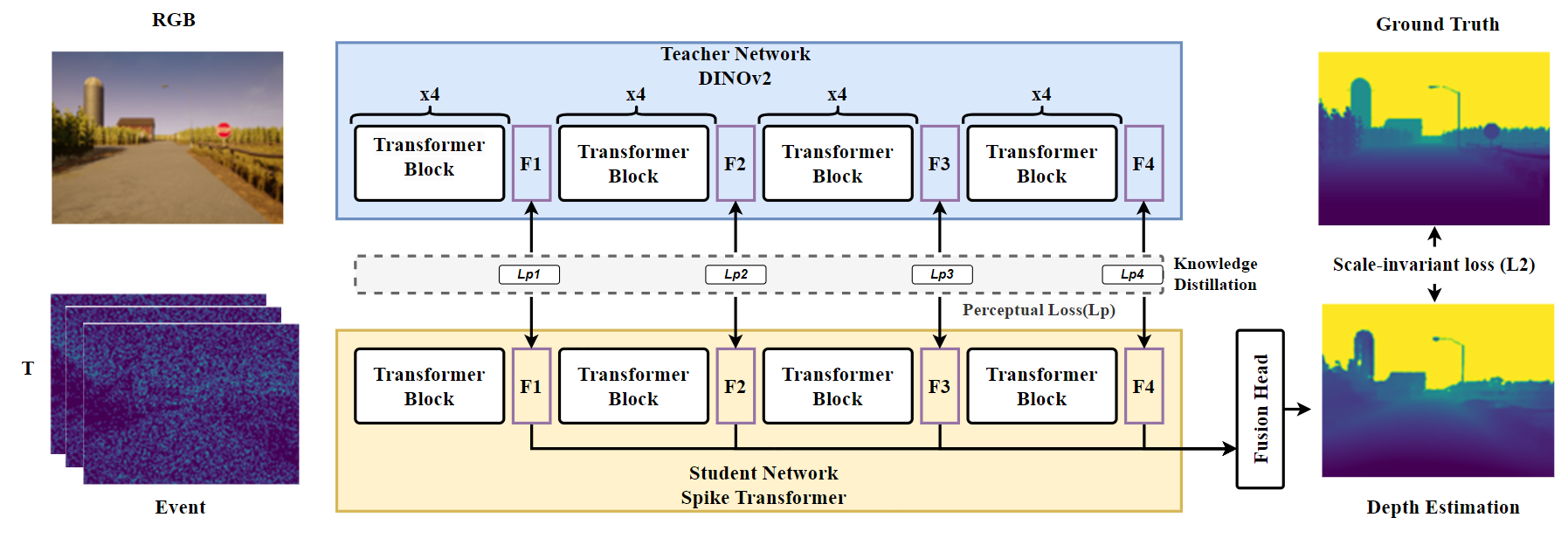}
\caption{The flowchart of the proposed method}
\label{fig:1}
\end{figure*}

We propose a novel energy-efficient spike transformer network for depth estimation via cross-modality knowledge distillation. The flowchart of the method is illustrated in {Figure~\ref{fig:1}}, encompassing three primary components:  1) Spike-driven transformer, 2) Fusion depth estimation head, and 3) Knowledge distillation. 
The rationale for our proposed method centres on three key innovations:

1)	We introduce a pure spike-driven transformer network that eliminates conventional floating-point operations. While transformers excel at capturing long-range dependencies, their matrix multiplication operations are computationally intensive. Our spike-driven design using binary spikes instead of floating-point values significantly reduces energy consumption while maintaining high performance through carefully designed spike-based attention and residual mechanisms.

2)	We propose a novel fusion depth estimation head designed to integrate features from multiple transformer stages for precise and robust depth estimation. Compared to existing methods, our approach overcomes the limitations of CNN based architectures, which often lose critical spatial information due to downsampling. By leveraging transformers’ ability to retain dimensional consistency and integrating features at multiple levels, the fusion head achieves superior depth estimation accuracy. Additionally, it is fully compatible with spike-based computation models, making it both efficient and biologically plausible, providing a significant advantage for real-world applications requiring precision and robustness in challenging environments.

3)	To address the limited training data available for SNNs, we leverage knowledge from DINOv2, a large vision foundation model, through a novel single-stage cross-modality distillation framework. Rather than requiring separate training phases or an additional teacher model, our approach directly transfers relevant features from RGB to event data domains, enabling efficient training while preserving the spike-based computation paradigm.

\subsection{Spike-driven transformer}

The proposed spike transformer aligns with the foundational structure of the original Vision Transformer (ViT), encompassing a Spiking Patch Embedding and Spiking Transformer Block. Given an event sequence, \( I\in\mathbb{R}^{T\times C\times H\times W}\), the spike patch embedding is used to convert the input into a sequence of tokens that can be processed by the transformer architecture, where the event input is projected as spike-form patches \( X\in\mathbb{R}^{T\times N\times D}, N =\frac{H}{8}\times\frac{W}{8}\). Then, the spiking patches \( X\) are passed to the multi spiking transformer blocks (L). Considering that we have used knowledge distillation from the large model, this method uses only a minimum number of blocks as $\mathrm{L}=4$. Inspired by \citep{Zhou2022Spikformer:,Zhou2023Spikingformer:}, in order to avoid non-spike computations in traditional deep learning architectures, a Spiking Self Attention (SSA) and a Spiking MLP block are used in spiking transformer blocks.

\subsubsection{Spiking patch embedding}

\begin{figure}[!htbp]
\centering
\includegraphics[width=0.8\textwidth]{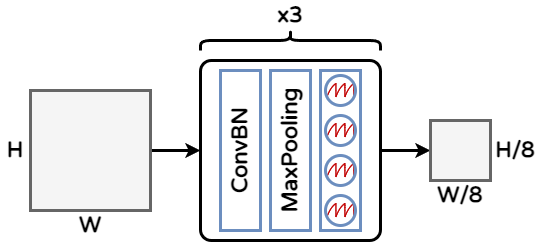}
\caption{The structure of spiking patch embedding}
\label{fig:2}
\end{figure}

In the original ViT  \citep{Dosovitskiy2020Image}, the patch embedding is used to represent an image as a sequence of tokens. This is done by dividing the image into a grid of patches and flattening each patch into a vector.  In this work, we implement this operation through a convolution batch norm (ConvBN), Max pooling (MP) and multistep LIF (MLIF) combination. The structure is shown in {Figure~\ref{fig:2}}. This process can be formulated as:

\begin{equation}
I_i=\operatorname{MLIF}(\operatorname{MP}(\operatorname{ConvBN}(I)))
\end{equation}

where the \( ConvBN\) contains 2D convolution layers (stride-1, 3 $\times$ 3 kernel size) and max-pooling. The number of operations can be bigger than 1. When multiple blocks are used, the number of output channels gradually increases and the size of the feature is halved, eventually matching the embedding dimension of the patch in ViT. 

\subsubsection{Spiking Transformer Block}

\begin{figure}[!htbp]
\includegraphics[width=0.9\textwidth]{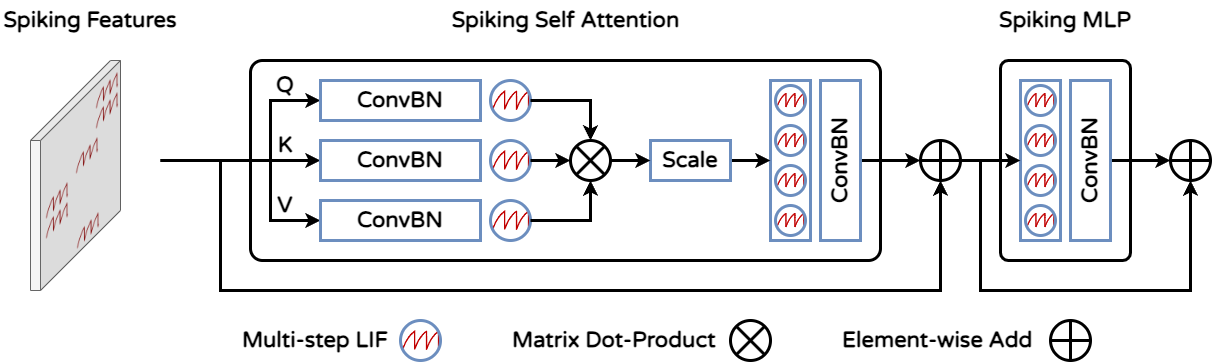}
\caption{The structure of transformer block}
\label{fig:3}
\end{figure}

The Spiking Transformer Block is structured to incorporate both a Spiking Self Attention (SSA) mechanism and a Spiking MLP block, as illustrated in {Figure~\ref{fig:3}}.

Guided by the findings in  \citep{Zhou2023Spikingformer:}, we position a multistep LIF before the ConvBN within the residual mechanism to omit floating-point multiplication and mixed-precision calculations during the ConvBN operation. This adjustment also enables ConvBN to replace conventional linear layers and batch normalization seamlessly. The SSA operation can be mathematically described as:

\begin{equation}
 \begin{gathered}
Q=\operatorname{MLIF}_Q\left(\operatorname{ConvBN}_Q\left(X^{\prime}\right)\right) \\
K=\operatorname{MLIF}_K\left(\operatorname{ConvBN}_K\left(X^{\prime}\right)\right) \\
V=\operatorname{MLIF}_V\left(\operatorname{ConvBN}_V\left(X^{\prime}\right)\right) \\
\operatorname{SSA}(Q, K, V)=\operatorname{ConvBN}\left(\operatorname{MLIF}\left(Q K^{\mathrm{T}} V * s\right)\right)
\end{gathered}
\end{equation}

The variables \( Q,K,V\in\mathbb{R}^{T\times N\times D}\) represent pure spike data (only containing 0 and 1).The Q, K, and V matrices are first obtained through learnable transformation matrices. These matrices are then converted into spiking sequences using distinct spiking neuron layers. The SSA leverages the inherently non-negative properties of the spike-form Q and K to produce a non-negative attention map. This makes softmax redundant as SSA can directly aggregate relevant features while ignoring irrelevant ones. The Scaling factor, \textit{s} is used to adjust the largest value of the matrix multiplication result. It does not affect the property of SSA. The Spiking MLP block consists of a residual connection and a combination of MLIF and ConvBN.

\subsection{Fusion Depth estimation Head}

The task of depth estimation requires generating pixel-wise depth predictions from encoded features. Two common approaches exist: the Fully Convolutional Network (FCN) head and fusion-based architectures. The FCN head applies convolution operations directly to the final encoder features to predict depths. While computationally efficient, this approach relies solely on high-level features and often struggles to preserve fine spatial details. 

Unlike CNN architectures that naturally produce multiscale feature hierarchies through downsampling, transformer networks maintain constant feature dimensions across layers. This poses a unique challenge for depth estimation, which benefits from both coarse semantic information and fine spatial details. Previous transformer-based segmentation models have addressed this through architectural modifications like the Multi-Scale Vision Transformer \citep{Fan2021Multiscale}  or Pyramid Vision Transformer \citep{Wang2021Pyramid}. However, such modifications could reduce the effectiveness of knowledge distillation from pretrained vision models because they are not modified with the same structure.

Therefore, we propose a fusion depth estimation head that leverages features from multiple transformer stages while preserving the original transformer architecture. The fusion head first projects the internal features from each transformer block into image-like representations. These multi-level features are then progressively combined through skip connections and upsampling operations to generate the final dense prediction.

\begin{figure*}[!htbp]
\centering
\includegraphics[width=0.9\textwidth]{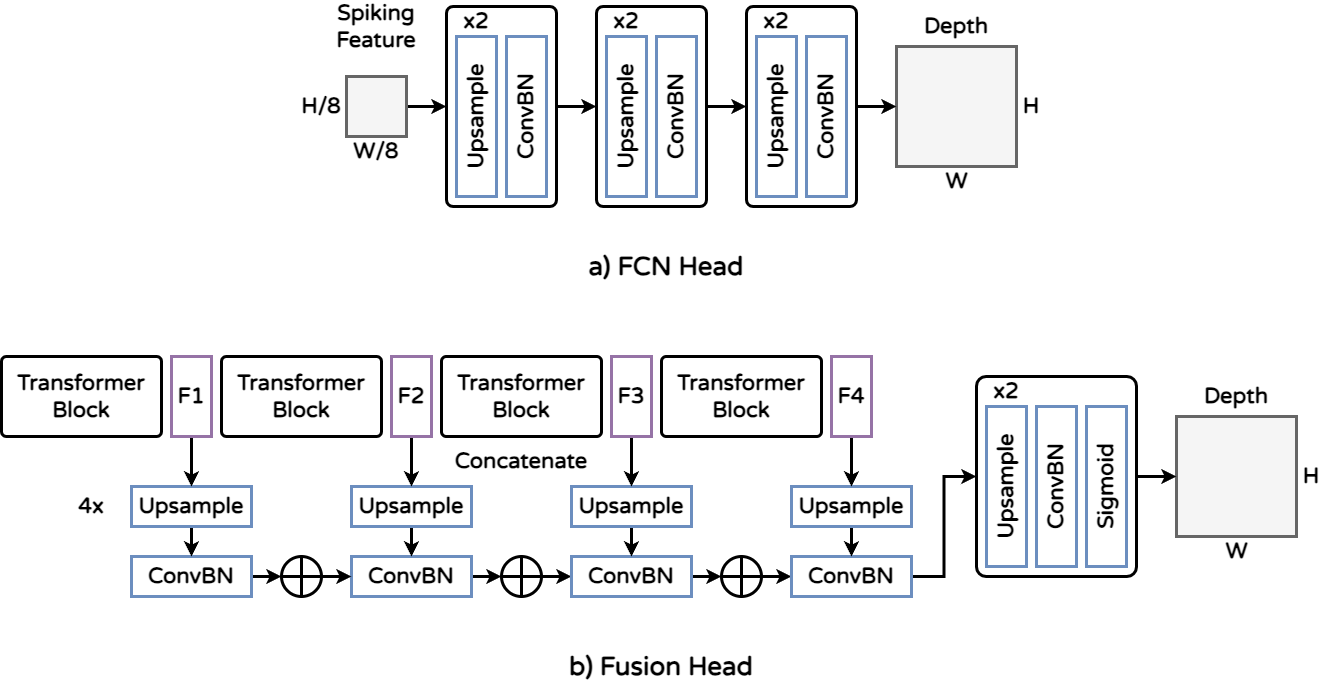}
\caption{The structure of the fusion head for depth estimation.}
\label{fig:4}
\end{figure*}

The structure of the fusion head for depth estimation is shown in {Figure~\ref{fig:4}}.

The first step of the fusion head is to assemble the internal features in transformer blocks into image-like feature representations. The feature representations are then fused into the final dense prediction with skip connections. A generic up sample structure is used to restore the feature representations to original data size. Give an input feature as \( F_{i}\in\mathbb{R}^{T\times H/8\times W/8}\)), i$=$1,2,3,4. The depth estimation head can be formulated as follows: 

\begin{equation}
\begin{gathered}\mathrm{Y}_2=\left(\operatorname{ConvBn}\left(\operatorname{Up}\left(F_1\right)\right)+\operatorname{Up}\left(F_2\right)\right) \\ \mathrm{Y}_3=\left(\operatorname{ConvBn}\left(\operatorname{Up}\left(\mathrm{Y}_2\right)\right)+\operatorname{Up}\left(F_3\right)\right) \\ \mathrm{Y}_4=\left(\operatorname{ConvBn}\left(\operatorname{Up}\left(\mathrm{Y}_3\right)\right)+\operatorname{Up}\left(F_4\right)\right) \\ Y=\operatorname{Sigmod}\left(\mathrm{Y}_4\right)\end{gathered}
\end{equation}
This design enables the network to combine high-level semantic information from deeper layers with fine-grained spatial details from earlier layers, leading to more accurate depth predictions while maintaining compatibility with knowledge distillation from vision foundation models.

\subsection{Knowledge distillation}

\begin{figure}[!htbp]
\centering
\includegraphics[width=0.9\textwidth]{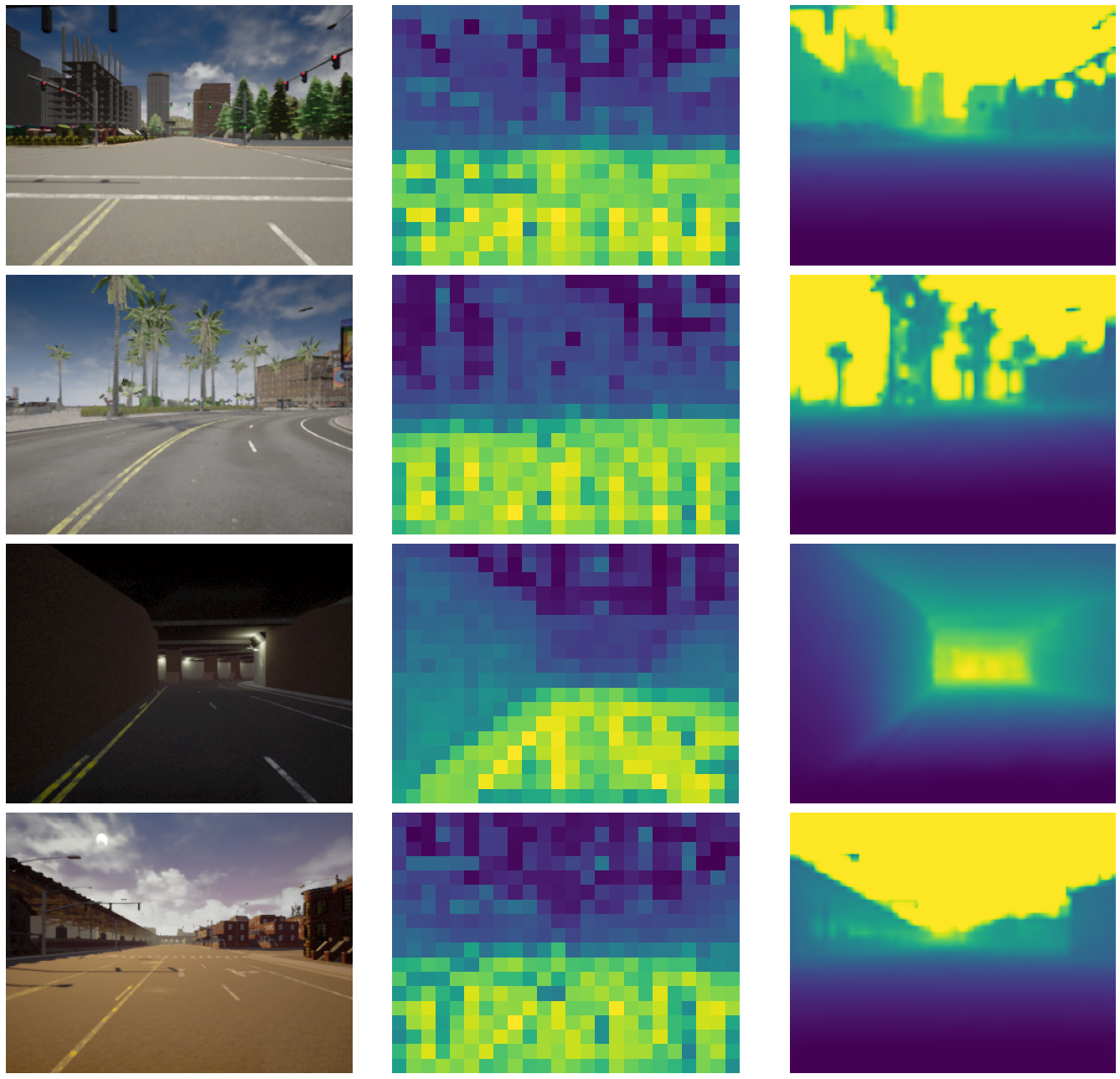}
\caption{a) The RGB image; b) Visualization of self-attention on DINOv2 features; (c) Depth estimation results using a linear probe on frozen DINOv2 features.}
\label{fig:5}
\end{figure}

Knowledge distillation is particularly challenging for SNNs due to two main factors: (1) the binary nature of spike data differs fundamentally from the continuous values used in traditional ANNs, and (2) the limited availability of labeled event camera data makes training challenging. To address these challenges, we propose a single-stage cross-modality knowledge distillation framework that leverages DINOv2 \citep{Oquab2023DINOv2:}, a large-scale vision foundation model, to guide our SNN training.

Our choice of DINOv2 as the teacher model is motivated by several key advantages:

1) Architectural Compatibility: DINOv2's Vision Transformer (ViT) architecture closely aligns with our model's structure, facilitating effective knowledge transfer due to their similar feature representations and computational patterns.

2) Rich Feature Representations: Pre-trained on 142 million diverse images, DINOv2 has demonstrated state-of-the-art performance in depth estimation tasks on benchmark datasets like NYU and SUN RGB-D. As shown in {Figure~\ref{fig:5}}, DINOv2's self-attention patterns and depth estimation capabilities on our dataset suggest it can provide valuable guidance during the knowledge distillation process.

3) Zero-shot Generalization: DINOv2's strong zero-shot learning capabilities enable effective knowledge transfer even when dealing with limited event camera data.

The knowledge distillation process can be shown in {Figure~\ref{fig:1}}. We freeze the DINOv2 (Lightblue) as a teacher model. The output features from DINOv2 are considered as targets in our training. To ensure compatible feature dimensions, we upsample the RGB input images by a factor of 1.75, resulting in teacher model features of size \( x_{rgb}^{'}\in\mathbb{R}^{d\times H/8\times W/8}\).

\begin{figure}[!htbp]
\includegraphics[width=0.9\textwidth]{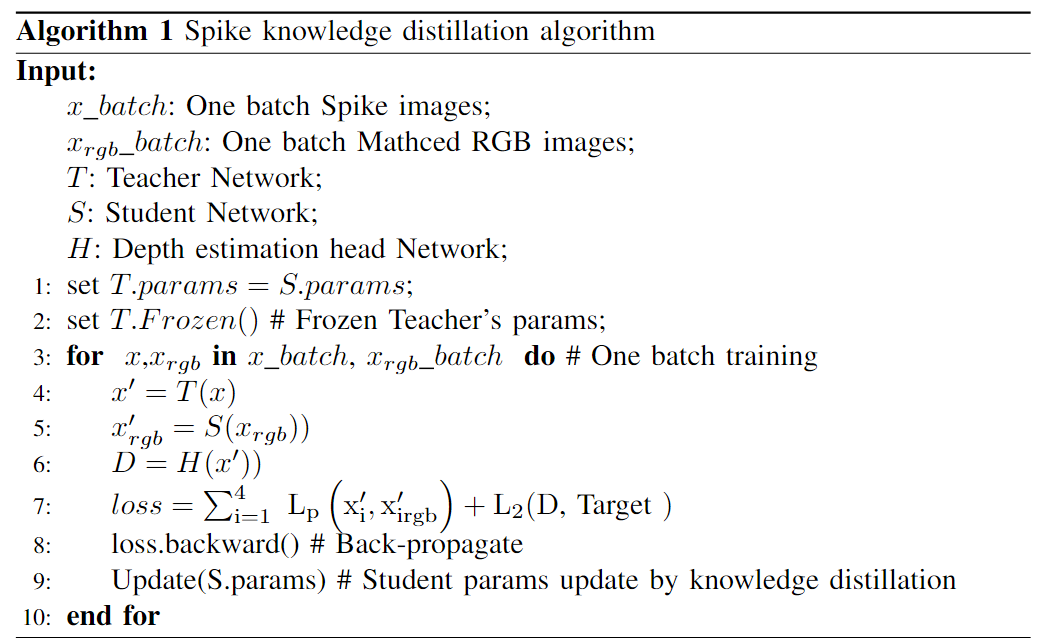}
\caption{The knowledge distillation algorithm}
\label{fig:6}
\end{figure}

{Figure~\ref{fig:6}} shows the knowledge distillation algorithm. Our distillation framework employs a fusion loss function that combines two complementary components:

Feature Perceptual Loss (Lp): Measures the distance between student and teacher feature representations, ensuring the SNN learns similar feature patterns. The Perceptual Loss \citep{Johnson2016Perceptual}  is used here to help capture high-level semantic differences between teacher and student representations, going beyond pixel-level comparisons.

The L2 loss function, a scale-invariant metric \citep{Eigen2015Predicting}, specifically designed for monocular depth estimation to address the inherent scale ambiguity problem.This scale-invariant loss is particularly important as it focuses on relative depth relationships rather than absolute values, aligning with the fundamental nature of monocular depth estimation where absolute scale cannot be determined from a single view. The combination of these loss terms enables our student network to learn both detailed features and consistent global depth relationships from the teacher model while maintaining computational efficiency through spike-based operations.

The fusion loss function is defined by the following equations:

\begin{equation}
\begin{gathered}
\mathcal{L}_{Pi}=\frac{1}{C \times H \times W}\left\|x_i-x_i^{\prime}\right\|_2^2 \\
\mathcal{L}_2=\frac{1}{n} \sum_{\mathbf{i}}\left(D_i^t-D_i^p\right)^2-\frac{1}{n^2}\left(\sum_{\mathbf{p}} D_i^t-D_i^p\right)^2
\end{gathered}
\end{equation}

Where \( D_{i}^{t}-D_{i}^{p}\)  is the difference between predicted and ground truth depth for pixel i,and n is total number of pixels with a dimension of \( H\times W\). It makes the loss invariant to uniform scaling of the depth predictions, allowing the network to learn consistent relative depth relationships even when absolute scale cannot be determined. This aligns with human depth perception, which relies heavily on relative rather than absolute depths.

\section{Experiments}

In this work, we conduct two experiments to demonstrate the effectiveness of the proposed SNN. We first introduce the details of datasets used in this experiment. Then, we evaluate the performance of our method, including accuracy and energy consumption. We evaluate our method on both real and synthetic event data to demonstrate the robustness and generalisability of the model. Finally, comprehensive ablation studies are conducted, which investigate the impact of each component.

\subsection{Datasets}

For model evaluation, we utilize two datasets comprising both real and synthetic data.

\textbf{DENSE Datasets:} The first dataset is a synthetic dataset from  \citep{Zhang2022Spike}, which is generated from DENSE dataset  \citep{Hidalgo-Carrio2020Learning}, including clear depth maps and intensity frames in 30 FPS under a variety of weather and illumination conditions. To obtain spike streams with high temporal resolution, the video is interpolated to generate intermediate RGB frames between adjacent 30-FPS frames. With absolute intensity information among RGB frames, each sensor pixel can continuously accumulate the light intensity with the spike generation mechanism, producing spike streams with a high temporal resolution (128$\times$30 FPS) that is 128 times of the video frame rate. The ‘spike’ version of DENSE dataset (namely DENSE spike) contains eight sequences, five for training, and three for evaluation. Each sequence consists of 999 samples, and each sample is a tuple of one RGB image, one depth map, and one spike stream. Each spike stream is simulated between two consecutive images, generating a binary sequence of 128 spike frames (with a size of 346 $\times$ 260 each) to depict the continuous process of dynamic scenes.  

\begin{figure}[!htbp]
\centering
\includegraphics[width=0.9\textwidth]{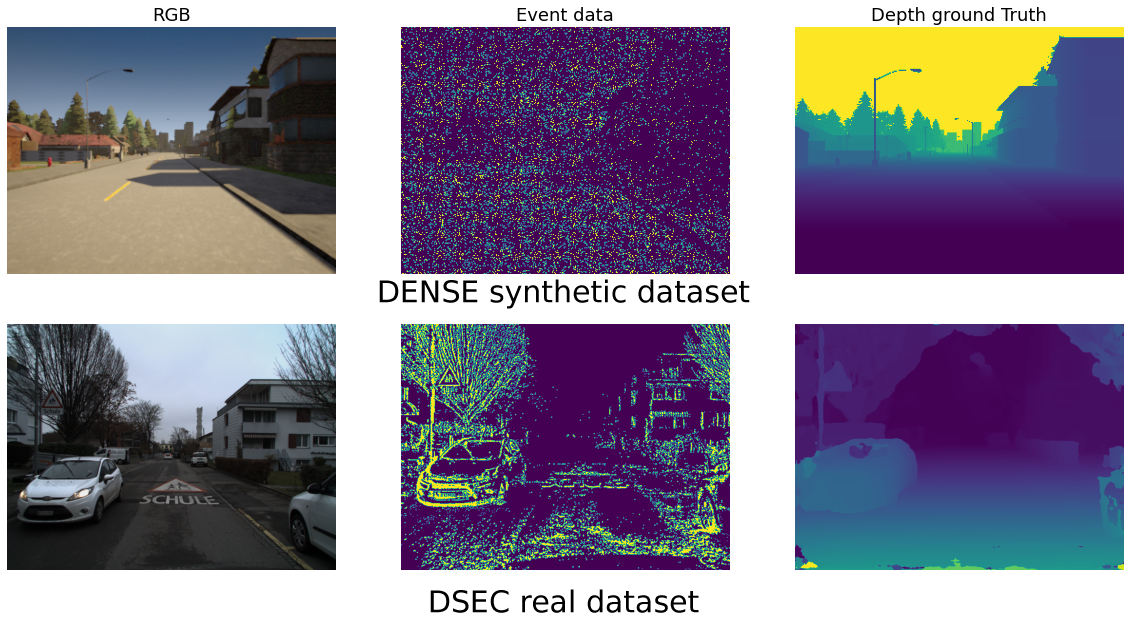}
\caption{Showcase of the two datasets.}
\label{fig:7}
\end{figure}

\textbf{DSEC Datasets:} The second dataset, DSEC, is a real event dataset that provides stereo dataset in driving scenarios. It contains data from two monochrome event cameras and two global shutter colour cameras in favourable and challenging illumination conditions. Hardware synchronised LiDAR data is also provided for depth prediction. The dataset contains 41 sequences collected by driving in a variety of illumination conditions and provides ground truth disparity for the depth estimation evaluation. In this work, 29 sequences (70$\%$) are used for model training and 12 are used for evaluation. Each sequence consists of 200-900 samples, and each sample is a tuple of one RGB image, one depth map (dense disparity), and one spike stream with 16 spike frames and size of 480 $\times$ 640. {Figure~\ref{fig:7}} presents the two data samples used in this work.

\subsection{Experiment design}

\subsubsection{Model performance}

In this section, we evaluate the depth estimation performance and energy consumption of our SNN on the synthetic (DENSE) and real datasets (DSEC) and compare it with three competing dense prediction networks, namely U-Net \citep{Ronneberger2015U-Net:}, E2Depth \citep{Hidalgo-Carrio2020Learning} and Spike-T \citep{Zhang2022Spike}. U-Net employs 2D convolutional layers as its encoder and focuses on spatial feature extraction, while E2Depth applies ConvLSTM layers that combine CNN and LSTM to capture the spatial and temporal features. The Spike-T employs transformer-based blocks to learn the spatio-temporal features simultaneously. These models therefore constitute our immediate and direct competitors.

\subsubsection{Ablation study}

An ablation study is detailed in this subsection to investigate the contributions of two novel components within our model. They are the fusion depth estimation head and the knowledge distillation technique. Their respective impacts on model performance are dissected and discussed.

\subsection{Metrics}

Several metrics are selected to evaluate the performance of the proposed method, including absolute relative error (Abs Rel.), square relative error (Sq Rel.), mean absolute depth error (MAE), root mean square logarithmic error (RMSE log) and the accuracy metric (Acc.$\delta$). The formulations are as follows:

\textbf{Absolute Relative Error (Abs Rel.)} computes average errors on the normalized depth map for every pixel, formulated as:

\begin{equation}
Abs Rel.=\frac{1}{N} \sum_p \frac{\left|\mathcal{D}_p-\widehat{\mathcal{D}}_p\right|}{\left|\mathcal{D}_p\right|}
\end{equation}

It normalises the value of depth to the range [0,1].

\textbf{Square Relative Error (Sq Rel.)}, formulated as 

\begin{equation}
Sq Rel.=\frac{1}{N} \sum_p \frac{\left|\mathcal{D}_p-\widehat{\mathcal{D}}_p\right|^2}{\left|\mathcal{D}_p\right|}
\end{equation}

which focuses on large depth errors due to its square numerator.

\textbf{Mean Absolute Error (MAE)} can be formulated as:

\begin{equation}
MAE=\frac{1}{N} \sum_p\left|\mathcal{D}_p-\widehat{\mathcal{D}}_p\right|
\end{equation}

\textbf{Root Mean Square Error (RMSE)} is a classic metric for per-pixel prediction error and the logarithm version can be denoted as

\begin{equation}
RMSE=\sqrt{\frac{1}{N} \sum_p\left|\log \mathcal{D}_p-\log \widehat{\mathcal{D}}_p\right|^2}
\end{equation}

\textbf{The Accuracy (Acc)} as $\delta$ denotes the percentage of all pixels \(\mathcal{D}_{p}\) that satisfy max: 

\begin{equation}
Acc=\left(\frac{\widehat{\mathcal{D}}_p}{\mathcal{D}_p}, \frac{\mathcal{D}_p}{\widehat{\mathcal{D}}_p}\right)<t h r
\end{equation}

where \( thr\) $=$ 1.25, 1.25\textsuperscript{2}, 1.25 \citep{Chen2019Live}.

\subsection{Experiment Result}

\subsubsection{Model performances}

Our experimental investigation encompasses both quantitative performance and energy consumption analyses, utilizing synthetic (DENSE) and real-world (DESC) datasets. Several metrics were employed to evaluate the outcomes comprehensively.

\begin{table*}[!htbp]
\begin{adjustbox}{max width=\textwidth}
\begin{tabular}{lllllllllll}
\hline
         & Abs   Rel ↓ & Sq   Rel ↓ & RMS   log ↓ & SI   log ↓ & $\delta < 1.25$ ↑ & $\delta < 1.25^2$ ↑ & $\delta < 1.25^3$ ↑ & Spike-driven & Param   (M) & Power   (mJ) \\ \hline
U-Net    & 2.89        & 72.25      & 0.19        & 1.73       & 0.39                 & 0.50                  & 0.58                  & No              & 31.20       & 72.93        \\
E2Depth  & 9.91        & 96.01      & 0.30        & 1.70       & 0.21                 & 0.31                  & 0.45                  & No              & 10.71       & 59.25        \\
Spike-T  & 1.57        & 39.77      & 0.17        & 0.91       & 0.50                 & 0.65                  & 0.74                  & No              & 35.68       & 41.77        \\
Proposed & 0.80        & 8.32       & 0.17        & 0.46       & 0.53                 & 0.68                  & 0.76                  & Yes             & 20.55       & 12.43        \\ \hline
\end{tabular}
\end{adjustbox}
\caption{Quantitative performance comparison on the synthetic datasets (DENSE) using various models. Results for both the validation set and the test set are presented. Symbols $\downarrow$ and $\uparrow$ indicate that a lower value and higher value are preferable, respectively. The "Param" column refers to the number of parameters, and "Power" provides the average theoretical energy consumption for predicting an image.}
\label{tab:1}
\end{table*}

Table~\ref{tab:1} presents a quantitative performance comparison using synthetic datasets (specifically the DENSE dataset).The results consistently indicate that the proposed method outperforms the alternative approaches across nearly all evaluated metrics.

Notably, substantial reductions are observed in the absolute relative error (Abs.Rel) and squared relative error (Sq.Rel)—critical metrics in depth estimation tasks. The Abs.Rel for the proposed method, recorded at 0.80, represents reductions of 72\%, 91.9\%, and 49\% compared to U-Net (2.89), E2Depth (9.91), and Spike-T (1.57), respectively. Similarly, the Sq.Rel of our method, at 8.32, demonstrates reductions of 88.5\%, 91.3\%, and 79\% relative to U-Net (72.25), E2Depth (96.01), and Spike-T (39.77).

In addition to error reduction, the proposed method achieves modest increases in accuracy metrics ($\delta < 1.25$, $\delta < 1.25^2$, and $\delta < 1.25^3$), attaining values of 0.53, 0.68, and 0.76, respectively. These values are slightly higher than those achieved by the competing methods, reinforcing the method’s superior capability in depth estimation tasks.

In terms of power consumption, the proposed method, which leverages pure spike computing, shows a marked advantage over its competitors, reducing power consumption by up to 82.9\% compared to U-Net (from 72.93mJ to 12.43mJ). Furthermore, the adoption of knowledge distillation has facilitated the use of merely four Transformer blocks in our method, significantly reducing the parameters by 42.4\% compared to the Spike-T method (from 35.68M to 20.55M), which utilizes eight Transformer blocks.

\begin{figure*}[!htbp]
\centering
\includegraphics[width=0.9\textwidth]{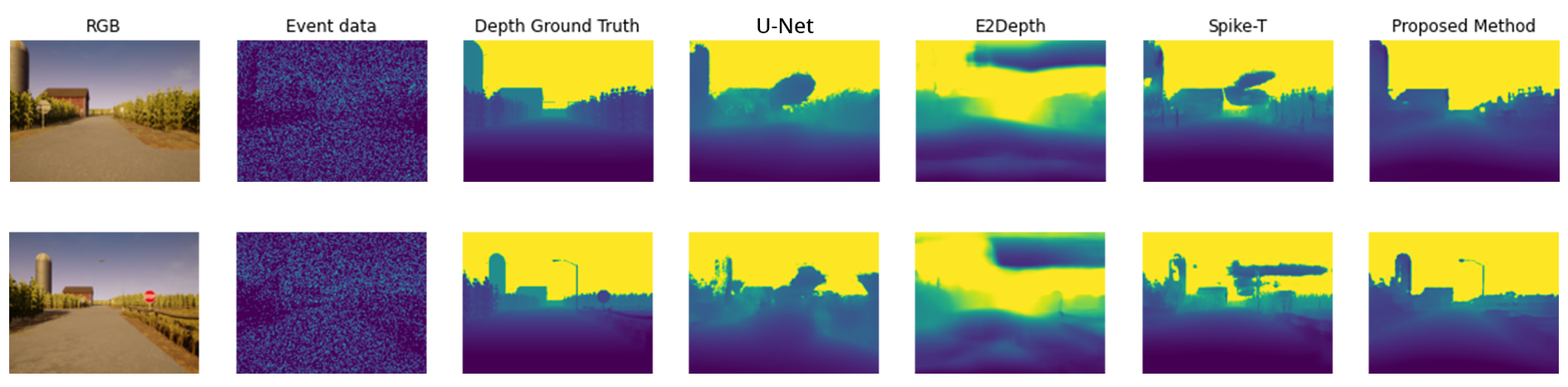}
\caption{Visualization results of multiple comparison models on the synthetic (DENSE) dataset.}
\label{fig:8}
\end{figure*}

These experimental results show that our proposed method can more effectively capture the spatial-temporal characteristics of irregular continuous spike data streams, delivering satisfactory accuracy. This is further illustrated in {Figure~\ref{fig:8}}, which depicts the visualization results from multiple comparison models on a validation synthetic dataset. The visualization demonstrates that, unlike the U-Net and Spike-T methods which can predict details yet misestimate depth, or the E2Depth method that produces blurry outcomes losing fine details, our method effectively manages to capture more intricate details, including minute structures, sharp edges, and contours.

In addition, in order to validate the generalisability of the model, we evaluate the proposed model on DESC real event dataset and compare it with the three competing models (E2Depth, Spike-T and U-Net). It is noteworthy that while the DENSE synthetic dataset encompasses 128 spike frames, the real-world DESC dataset contains only 16 frames. Our model and the SpikeU-Net model require retraining on this reduced dataset. However, the SNN-based methods (E2Depth and Spike-T) are unable to be retrained due to insufficient training parameters, necessitating the replication of DESC data to 128 frames to accommodate their setups. Consequently, the performance of E2Depth and Spike-T are expectedly low.

The results, as shown in {Table~\ref{tab:2}}, demonstrated that the proposed model excels across all metrics in comparison to the E2Depth, Spike-T, and U-Net models. This superior performance is evident particularly in terms of metrics such as Abs Rel, Sq Rel, RMS log, and SI log, as well as in the accuracy metrics ($\delta < 1.25$, $\delta < 1.25^2$, and $\delta < 1.25^3$), where higher scores are indicative of better performance. These findings underscore the effectiveness of the proposed model in handling real-event data from spiking cameras.

\begin{table}[!htbp]
\begin{adjustbox}{max width=0.9\textwidth}
\begin{tabular}{llllllll}
\hline
           & Abs Rel   ↓ & Sq Rel ↓ & RMS log   ↓ & SI log ↓ & $\delta < 1.25$ ↑ & $\delta < 1.25^2$ ↑ & $\delta < 1.25^3$ ↑ \\ \hline
E2Depth    & 9.909       & 96.011   & 0.299       & 1.697    & 0.209                & 0.309                 & 0.448                 \\
Spike-T    & 2.853       & 51.757   & 0.321       & 0.751    & 0.164                & 0.341                 & 0.483                 \\
SpikeU-Net & 1.203       & 3.281    & 0.181       & 1.210    & 0.142                & 0.309                 & 0.502                 \\
Proposed   & 1.000       & 0.999    & 0.105       & 0.212    & 0.387                & 0.500                 & 0.583                 \\ \hline
\end{tabular}
\end{adjustbox}

\caption{Quantitative comparison on the DESC real dataset.}
\label{tab:2}
\end{table}

\begin{figure*}[!htbp]
\centering
\includegraphics[width=0.9\textwidth]{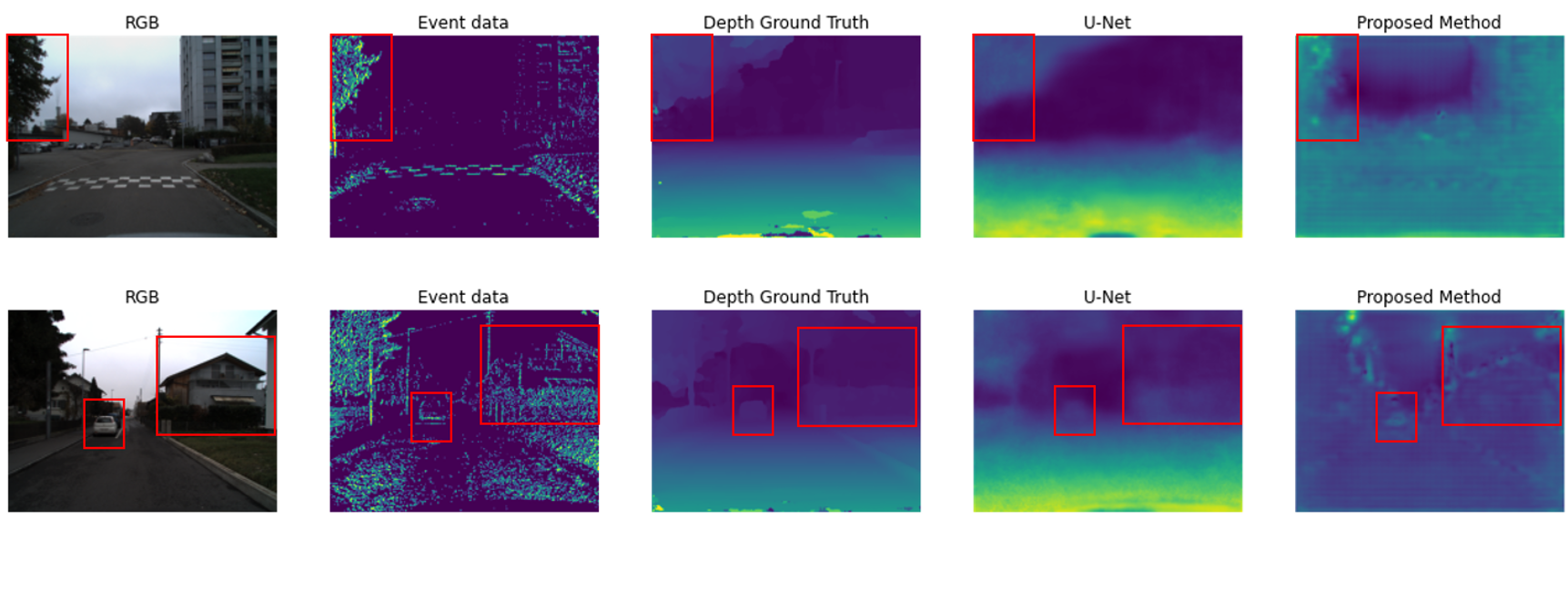}
\caption{Visualization results under low-light conditions on the real(DESC) dataset.}
\label{fig:9}
\end{figure*}

{Figure~\ref{fig:9}} shows the visualisation result in a low light environment. Our method effectively identifies features such as trees and houses along the roadside, as well as vehicles located in the center of the road.

\subsubsection{Ablation study}

This subsection presents an ablation study conducted to evaluate the effectiveness of the proposed Fusion Depth Estimation Head and Knowledge Distillation (KD) modules. 

\begin{table}[!htbp]
\begin{adjustbox}{max width=0.9\textwidth}
\begin{tabular}{llllllll}
\hline
                & Abs Rel ↓ & Sq Rel ↓ & RMS log ↓ & SI log ↓ & $\delta < 1.25$ ↑ & $\delta < 1.25^2$ ↑ & $\delta < 1.25^3$ ↑ \\ \hline
Linear FCN Head & 2.85      & 51.76    & 0.32      & 0.75     & 0.16               & 0.34                & 0.48                \\
W/O KD          & 2.89      & 72.25    & 0.19      & 1.73     & 0.39               & 0.50                & 0.58                \\
proposed        & 0.80      & 8.32     & 0.17      & 0.46     & 0.53               & 0.68                & 0.76                \\ \hline
\end{tabular}
\end{adjustbox}
\caption{Quantitative performance comparison on the synthetic (DENSE) dataset in the ablation study.}
\label{tab:3}\end{table}

\begin{figure*}[!htbp]
\centering
\includegraphics[width=0.9\textwidth]{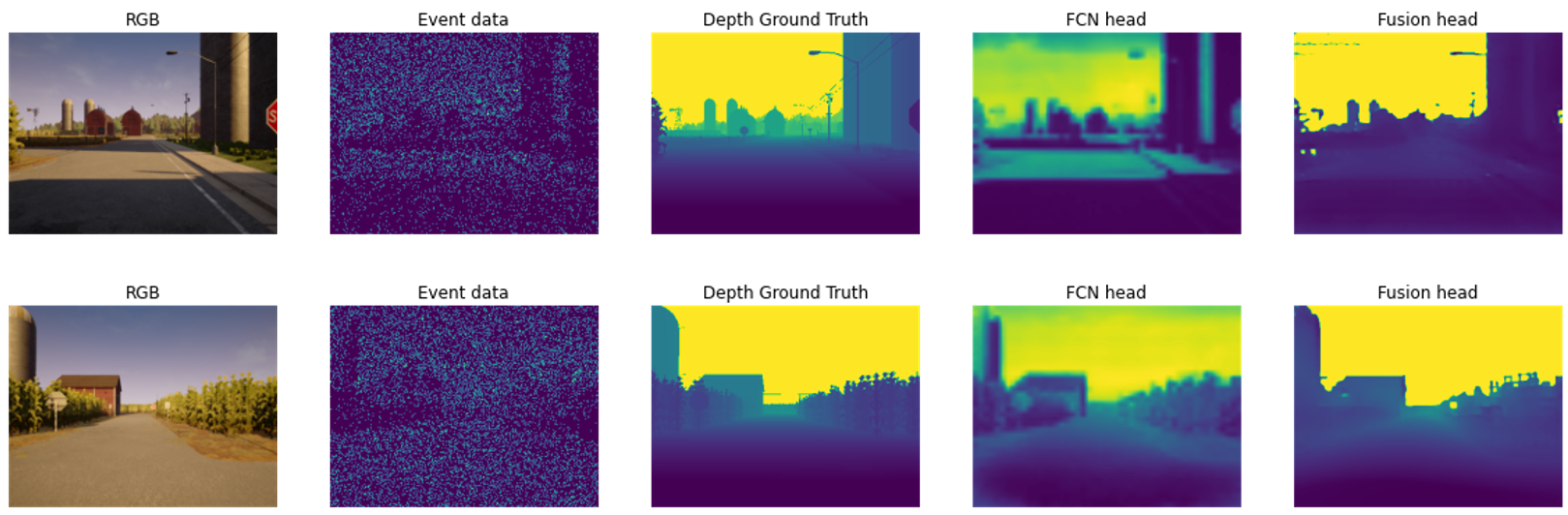}
\caption{Visualization results on validation synthetic (DENSE) dataset by using FCN head and proposed fusion depth estimation head.}
\label{fig:10}
\end{figure*}

{Table~\ref{tab:3}} reports the quantitative performance comparison on the synthetic datasets (DENSE) in the ablation study. As demonstrated by the results, all accuracy metrics exhibit a decline when employing the linear FCN (Fully Convolutional Network) head for depth estimation. Specifically, Absolute Relative (Abs Rel) error increased from 0.80 to 2.85, while the Squared Relative (Sq Rel) error escalated from 8.32 to 51.76. {Figure~\ref{fig:10}} illustrates the visualization results of using two different heads. The image becomes notably blurrier and loses details when employing the linear FCN head, which relies solely on the final features generated by the transformer. Conversely, our fusion head integrates multi-scale features, thereby facilitating superior recovery of details compared to the linear FCN head.

\begin{figure*}[!htbp]
\centering
\includegraphics[width=0.9\textwidth]{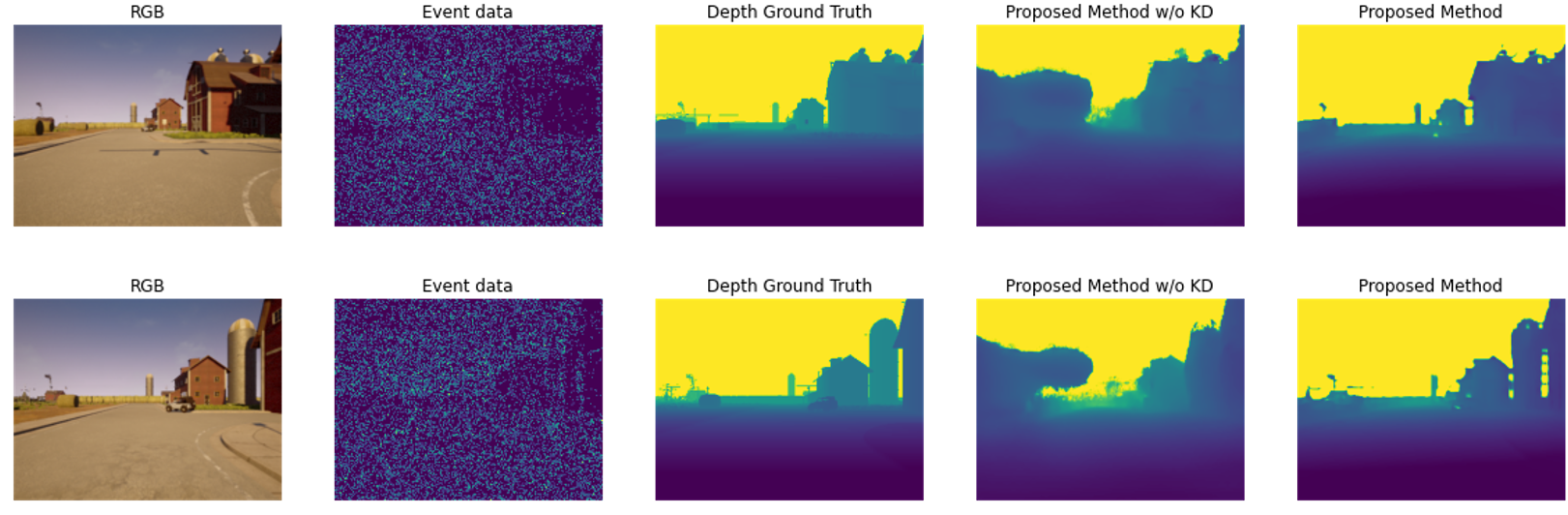}
\caption{Visualization results on validation synthetic (DENSE) dataset with and without knowledge distillation.}
\label{fig:11}
\end{figure*}

The visualization of results with and without knowledge distillation is depicted in {Figure~\ref{fig:11}}. The results without knowledge distillation approximate to those of the baseline models from Experiment 1. The presence of noise in the spike data leads to less accurate depth estimations for certain segments of the point cloud. However, employing knowledge distillation enables the model to predict the depths of distant clouds more accurately, a benefit attributed to the enhanced inductive capabilities derived from the substantial foundational model.

\section{Discussion}

The development of efficient and accurate depth estimation methods for event cameras remains a significant challenge in computer vision\citep{Gallego2022Event-Based, Tayarani-Najaran2021Event-Based}. Our work addresses this challenge through three key innovations: a purely spike-driven transformer architecture, fusion depth estimation head and a novel knowledge distillation framework. The experimental results demonstrate the improvements in both accuracy and energy efficiency, we discuses the implications, limitations, and potential impact of our approach in this section.

The satisfactory performance of the spike transformer architecture can be attributed to several factors. First, by implementing spike-driven residual learning and self-attention mechanisms \citep{Zhou2022Spikformer:, Zhou2023Spikingformer:}, our model effectively captures the temporal dynamics inherent in event camera data while maintaining the computational efficiency characteristic of SNNs \citep{Rafi2021brief}. This is evidenced by the significant reductions in absolute relative error (49\% reduction) and squared relative error (39.77\% reduction) compared to the state-of-the-art Spike-T model \citep{Zhang2022Spike}. The elimination of floating-point operations in the transformer portion substantially reduces power consumption \citep{Han2020deep, Lemaire2022analytical}, making our approach more practical for real-world applications where energy efficiency is crucial.


Our innovative fusion depth estimation head integrates multi-scale transformer features, preserving fine details and global structure, addressing limitations of traditional methods and ensuring robust performance. Traditional approaches often struggle to maintain fine-grained spatial information while processing temporal event data \citep{Ming2021Deep}. Our fusion head addresses this by effectively combining features from multiple transformer stages, as demonstrated by the improved preservation of detail in our depth predictions compared to standard linear FCN heads \citep{Ronneberger2015U-Net:}. This multi-scale feature integration is particularly beneficial for capturing both fine details and global scene structure \citep{Fan2021Multiscale}, contributing to the overall robustness of our depth estimates. However, it is important to acknowledge the limitation that the fusion depth estimation head is not purely spike-based.  This decision was driven by the requirement for high accuracy in depth estimation tasks. Pure spike-based operations, while more energy-efficient, currently face limitations for precision value prediction.

Our hybrid approach allows us to leverage the energy efficiency of spike-based computing in the feature extraction stages while maintaining the high accuracy requirements of depth estimation through conventional computation in the fusion head.

Our knowledge distillation framework represents a significant advancement in addressing the longstanding challenge of training SNNs with limited data \citep{He2023Improving}. By leveraging the rich feature representations learned by DINOv2 \citep{Oquab2023DINOv2:} from its vast training dataset, our approach effectively transfers knowledge across modalities (from traditional images to event data). Ablation studies clearly demonstrate the value of this knowledge transfer, showing marked improvements in depth estimation accuracy and robustness to noise when knowledge distillation is employed \citep{Gou2021Knowledge}. This is particularly evident in the model's ability to handle challenging scenarios such as estimating depths of distant objects and operating in low-light conditions \citep{Furmonas2022Analytical}.

Several avenues for future research emerge from our findings. First, investigating methods to achieve comparable accuracy with pure spike-based fusion mechanisms remains an important challenge, though this may require fundamental advances in spike-based computing precision. Second, evaluation on neuromorphic hardware platforms (such as SpiNNaker 1/2 \citep{Furber2014spinnaker,mayr2019spinnaker}, BrainScales \citep{Pehle2022brainscales}, or TrueNorth \citep{Lohr2020complex}) would provide valuable insights into real-world performance and energy efficiency.  Especially SpiNNaker 2\citep{mayr2019spinnaker}, unlike traditional methods, it is specifically designed to handle operations such as multiplications, which are generally inefficient for spiking computations. The introduction of such platforms highlights the necessity for hybrid approaches, where specific operations may leverage conventional hardware optimizations while retaining the efficiency of spike-driven designs.

Additionally, our architecture's ability to effectively process temporal event data suggests potential applications beyond depth estimation, such as object tracking \citep{Chen2019Live} or motion estimation \citep{Huang2022Real-time}. The success of our knowledge distillation approach also raises interesting questions about the broader applicability of foundation models in training efficient SNNs for various computer vision tasks \citep{Bommasani2022On}.

The integration of event cameras in autonomous systems and robotics applications continues to grow \citep{Cao2021Fusion-based}, driven by their advantages in terms of latency, dynamic range, and power efficiency \citep{Posch2011QVGA}. Our work demonstrates that by combining the biological inspiration of SNNs with modern deep learning architectures and knowledge distillation techniques, we can develop more efficient and accurate methods for processing event camera data. While our hybrid approach represents a careful balance between computational efficiency and accuracy requirements, it provides valuable insights for the broader development of neuromorphic computing systems \citep{Auge2021Survey}. The success of this approach suggests that future developments in spike-based computing may benefit from similar pragmatic trade-offs between pure neuromorphic computation and task-specific performance requirements.

\section{Conclusion}

In this paper, we have introduced a novel energy-efficient Spike Transformer network for depth estimation, leveraging spiking camera data. The proposed architecture integrates spike-driven residual learning and spiking self-attention mechanisms, creating a transformer framework that operates entirely within the spike domain. This innovative design achieves significant computational efficiency, with an 82.9\% reduction in power consumption compared to conventional methods (from 72.93mJ to 12.43mJ per inference). Additionally, our single-stage knowledge distillation framework, leveraging large foundational ANN models such as DINOv2, enables robust training of Spiking Neural Networks (SNNs) even in the presence of limited data.   Extensive evaluations on synthetic and real-world datasets demonstrate the efficacy of our approach, with significant improvements in key performance metrics, including a 49\% reduction in Absolute Relative Error and a 39.77\% reduction in Square Relative Error compared to the state-of-the-art SpikeT model. The architecture further enhances efficiency with a 42.4\% reduction in parameters (20.55M versus 35.68M), making it particularly well-suited for resource-constrained environments. By combining high accuracy with remarkably low power requirements, our spike-based design is well-suited for practical applications. Future work will focus on extending this research to broader real-world scenarios, including deployment on dedicated SNN processors and further validation with diverse datasets. These efforts aim to unlock the full potential of Spike Transformers in applications such as autonomous navigation, robotics, and energy-efficient vision systems, paving the way for advanced neuromorphic computing in practical settings .

\section*{Acknowledgement}

Funding acknowledgment: EPSRC (EP/X013707/1), BBSRC(BB/R019983/1, BB/S020969/1), Innovate UK(10091423).
 
\bibliographystyle{elsarticle-harv} 
\bibliography{spike}

\begin{thebibliography}{62}
\expandafter\ifx\csname natexlab\endcsname\relax\def\natexlab#1{#1}\fi
\providecommand{\url}[1]{\texttt{#1}}
\providecommand{\href}[2]{#2}
\providecommand{\path}[1]{#1}
\providecommand{\DOIprefix}{doi:}
\providecommand{\ArXivprefix}{arXiv:}
\providecommand{\URLprefix}{URL: }
\providecommand{\Pubmedprefix}{pmid:}
\providecommand{\doi}[1]{\href{http://dx.doi.org/#1}{\path{#1}}}
\providecommand{\Pubmed}[1]{\href{pmid:#1}{\path{#1}}}
\providecommand{\bibinfo}[2]{#2}
\ifx\xfnm\relax \def\xfnm[#1]{\unskip,\space#1}\fi
\bibitem[{Arnab et~al.(2021)Arnab, Dehghani, Heigold, Sun, Lučić and Schmid}]{Arnab2021ViViT:}
\bibinfo{author}{Arnab, A.}, \bibinfo{author}{Dehghani, M.}, \bibinfo{author}{Heigold, G.}, \bibinfo{author}{Sun, C.}, \bibinfo{author}{Lučić, M.}, \bibinfo{author}{Schmid, C.}, \bibinfo{year}{2021}.
\newblock \bibinfo{title}{Vivit: A video vision transformer}, pp. \bibinfo{pages}{6836--6846}.
\newblock \bibinfo{note}{[Online; accessed 2023-08-15]}.
\bibitem[{Auge et~al.(2021)Auge, Hille, Mueller and Knoll}]{Auge2021Survey}
\bibinfo{author}{Auge, D.}, \bibinfo{author}{Hille, J.}, \bibinfo{author}{Mueller, E.}, \bibinfo{author}{Knoll, A.}, \bibinfo{year}{2021}.
\newblock \bibinfo{title}{A survey of encoding techniques for signal processing in spiking neural networks}.
\newblock \DOIprefix\doi{10.1007/s11063-021-10562-2}.
\bibitem[{Bommasani et~al.(2021)Bommasani, Hudson, Adeli, Altman, Arora, von Arx, Bernstein, Bohg, Bosselut, Brunskill et~al.}]{Bommasani2022On}
\bibinfo{author}{Bommasani, R.}, \bibinfo{author}{Hudson, D.A.}, \bibinfo{author}{Adeli, E.}, \bibinfo{author}{Altman, R.}, \bibinfo{author}{Arora, S.}, \bibinfo{author}{von Arx, S.}, \bibinfo{author}{Bernstein, M.S.}, \bibinfo{author}{Bohg, J.}, \bibinfo{author}{Bosselut, A.}, \bibinfo{author}{Brunskill, E.}, et~al., \bibinfo{year}{2021}.
\newblock \bibinfo{title}{On the opportunities and risks of foundation models}.
\bibitem[{Cao et~al.(2021)Cao, Chen, Xia, Zhuang and Knoll}]{Cao2021Fusion-based}
\bibinfo{author}{Cao, H.}, \bibinfo{author}{Chen, G.}, \bibinfo{author}{Xia, J.}, \bibinfo{author}{Zhuang, G.}, \bibinfo{author}{Knoll, A.}, \bibinfo{year}{2021}.
\newblock \bibinfo{title}{Fusion-based feature attention gate component for vehicle detection based on event camera}.
\newblock \bibinfo{note}{Publisher: IEEE}.
\bibitem[{Chen et~al.(2020)Chen, Cao, Conradt, Tang, Rohrbein and Knoll}]{Chen2020Event-Based}
\bibinfo{author}{Chen, G.}, \bibinfo{author}{Cao, H.}, \bibinfo{author}{Conradt, J.}, \bibinfo{author}{Tang, H.}, \bibinfo{author}{Rohrbein, F.}, \bibinfo{author}{Knoll, A.}, \bibinfo{year}{2020}.
\newblock \bibinfo{title}{Event-based neuromorphic vision for autonomous driving: A paradigm shift for bio-inspired visual sensing and perception}.
\newblock \DOIprefix\doi{10.1109/MSP.2020.2985815}. \bibinfo{note}{event-title: IEEE Signal Processing Magazine}.
\bibitem[{Chen and Guo(2019)}]{Chen2019Live}
\bibinfo{author}{Chen, S.}, \bibinfo{author}{Guo, M.}, \bibinfo{year}{2019}.
\newblock \bibinfo{title}{Live demonstration: Celex-v: A 1m pixel multi-mode event-based sensor}, pp. \bibinfo{pages}{1682--1683}.
\newblock \DOIprefix\doi{10.1109/CVPRW.2019.00214}. \bibinfo{note}{iSSN: 2160-7516}.
\bibitem[{Cordone et~al.(2021)Cordone, Miramond and Ferrante}]{Cordone2021Learning}
\bibinfo{author}{Cordone, L.}, \bibinfo{author}{Miramond, B.}, \bibinfo{author}{Ferrante, S.}, \bibinfo{year}{2021}.
\newblock \bibinfo{title}{Learning from event cameras with sparse spiking convolutional neural networks}.
\newblock \DOIprefix\doi{10.1109/IJCNN52387.2021.9533514}. \bibinfo{note}{iSSN: 2161-4407}.
\bibitem[{Diehl et~al.(2016)Diehl, Zarrella, Cassidy, Pedroni and Neftci}]{Diehl2016Conversion}
\bibinfo{author}{Diehl, P.U.}, \bibinfo{author}{Zarrella, G.}, \bibinfo{author}{Cassidy, A.}, \bibinfo{author}{Pedroni, B.U.}, \bibinfo{author}{Neftci, E.}, \bibinfo{year}{2016}.
\newblock \bibinfo{title}{Conversion of artificial recurrent neural networks to spiking neural networks for low-power neuromorphic hardware}, pp. \bibinfo{pages}{1--8}.
\newblock \DOIprefix\doi{10.1109/ICRC.2016.7738691}.
\bibitem[{Dosovitskiy et~al.(2020)Dosovitskiy, Beyer, Kolesnikov, Weissenborn, Zhai, Unterthiner, Dehghani, Minderer, Heigold, Gelly, Uszkoreit and Houlsby}]{Dosovitskiy2020Image}
\bibinfo{author}{Dosovitskiy, A.}, \bibinfo{author}{Beyer, L.}, \bibinfo{author}{Kolesnikov, A.}, \bibinfo{author}{Weissenborn, D.}, \bibinfo{author}{Zhai, X.}, \bibinfo{author}{Unterthiner, T.}, \bibinfo{author}{Dehghani, M.}, \bibinfo{author}{Minderer, M.}, \bibinfo{author}{Heigold, G.}, \bibinfo{author}{Gelly, S.}, \bibinfo{author}{Uszkoreit, J.}, \bibinfo{author}{Houlsby, N.}, \bibinfo{year}{2020}.
\newblock \bibinfo{title}{An image is worth 16x16 words: Transformers for image recognition at scale}.
\newblock \bibinfo{note}{ArXiv: 2010.11929}.
\bibitem[{Eigen and Fergus(2015)}]{Eigen2015Predicting}
\bibinfo{author}{Eigen, D.}, \bibinfo{author}{Fergus, R.}, \bibinfo{year}{2015}.
\newblock \bibinfo{title}{Predicting depth, surface normals and semantic labels with a common multi-scale convolutional architecture}.
\newblock \DOIprefix\doi{10.48550/arXiv.1411.4734}. \bibinfo{note}{arXiv:1411.4734 [cs]}.
\bibitem[{Fan et~al.(2021)Fan, Xiong, Mangalam, Li, Yan, Malik and Feichtenhofer}]{Fan2021Multiscale}
\bibinfo{author}{Fan, H.}, \bibinfo{author}{Xiong, B.}, \bibinfo{author}{Mangalam, K.}, \bibinfo{author}{Li, Y.}, \bibinfo{author}{Yan, Z.}, \bibinfo{author}{Malik, J.}, \bibinfo{author}{Feichtenhofer, C.}, \bibinfo{year}{2021}.
\newblock \bibinfo{title}{Multiscale vision transformers}.
\newblock \DOIprefix\doi{10.48550/arXiv.2104.11227}. \bibinfo{note}{arXiv:2104.11227 [cs]}.
\bibitem[{Fang et~al.(2021)Fang, Yu, Chen, Huang, Masquelier and Tian}]{Fang2021Deep}
\bibinfo{author}{Fang, W.}, \bibinfo{author}{Yu, Z.}, \bibinfo{author}{Chen, Y.}, \bibinfo{author}{Huang, T.}, \bibinfo{author}{Masquelier, T.}, \bibinfo{author}{Tian, Y.}, \bibinfo{year}{2021}.
\newblock \bibinfo{title}{Deep residual learning in spiking neural networks}.
\bibitem[{Furber et~al.(2014)Furber, Galluppi, Temple and Plana}]{Furber2014spinnaker}
\bibinfo{author}{Furber, S.B.}, \bibinfo{author}{Galluppi, F.}, \bibinfo{author}{Temple, S.}, \bibinfo{author}{Plana, L.A.}, \bibinfo{year}{2014}.
\newblock \bibinfo{title}{The spinnaker project}.
\bibitem[{Furmonas et~al.(2022)Furmonas, Liobe and Barzdenas}]{Furmonas2022Analytical}
\bibinfo{author}{Furmonas, J.}, \bibinfo{author}{Liobe, J.}, \bibinfo{author}{Barzdenas, V.}, \bibinfo{year}{2022}.
\newblock \bibinfo{title}{Analytical review of event-based camera depth estimation methods and systems}.
\newblock \bibinfo{note}{Publisher: MDPI}.
\bibitem[{Gallego et~al.(2022)Gallego, Delbrück, Orchard, Bartolozzi, Taba, Censi, Leutenegger, Davison, Conradt, Daniilidis and Scaramuzza}]{Gallego2022Event-Based}
\bibinfo{author}{Gallego, G.}, \bibinfo{author}{Delbrück, T.}, \bibinfo{author}{Orchard, G.}, \bibinfo{author}{Bartolozzi, C.}, \bibinfo{author}{Taba, B.}, \bibinfo{author}{Censi, A.}, \bibinfo{author}{Leutenegger, S.}, \bibinfo{author}{Davison, A.J.}, \bibinfo{author}{Conradt, J.}, \bibinfo{author}{Daniilidis, K.}, \bibinfo{author}{Scaramuzza, D.}, \bibinfo{year}{2022}.
\newblock \bibinfo{title}{Event-based vision: A survey}.
\newblock \DOIprefix\doi{10.1109/TPAMI.2020.3008413}. \bibinfo{note}{event-title: IEEE Transactions on Pattern Analysis and Machine Intelligence}.
\bibitem[{Gou et~al.(2021)Gou, Yu, Maybank and Tao}]{Gou2021Knowledge}
\bibinfo{author}{Gou, J.}, \bibinfo{author}{Yu, B.}, \bibinfo{author}{Maybank, S.J.}, \bibinfo{author}{Tao, D.}, \bibinfo{year}{2021}.
\newblock \bibinfo{title}{Knowledge distillation: A survey}.
\newblock \bibinfo{note}{ArXiv: 2006.05525}.
\bibitem[{Han and Roy(2020)}]{Han2020deep}
\bibinfo{author}{Han, B.}, \bibinfo{author}{Roy, K.}, \bibinfo{year}{2020}.
\newblock \bibinfo{title}{Deep spiking neural network: Energy efficiency through time based coding}.
\bibitem[{Han et~al.(2023)Han, Wang, Chen, Chen, Guo, Liu, Tang, Xiao, Xu, Xu, Yang, Zhang and Tao}]{Han2023Survey}
\bibinfo{author}{Han, K.}, \bibinfo{author}{Wang, Y.}, \bibinfo{author}{Chen, H.}, \bibinfo{author}{Chen, X.}, \bibinfo{author}{Guo, J.}, \bibinfo{author}{Liu, Z.}, \bibinfo{author}{Tang, Y.}, \bibinfo{author}{Xiao, A.}, \bibinfo{author}{Xu, C.}, \bibinfo{author}{Xu, Y.}, \bibinfo{author}{Yang, Z.}, \bibinfo{author}{Zhang, Y.}, \bibinfo{author}{Tao, D.}, \bibinfo{year}{2023}.
\newblock \bibinfo{title}{A survey on vision transformer}.
\newblock \DOIprefix\doi{10.1109/TPAMI.2022.3152247}. \bibinfo{note}{event-title: IEEE Transactions on Pattern Analysis and Machine Intelligence}.
\bibitem[{He et~al.(2023)He, Zhao, Li, Shen, Kong and Zeng}]{He2023Improving}
\bibinfo{author}{He, X.}, \bibinfo{author}{Zhao, D.}, \bibinfo{author}{Li, Y.}, \bibinfo{author}{Shen, G.}, \bibinfo{author}{Kong, Q.}, \bibinfo{author}{Zeng, Y.}, \bibinfo{year}{2023}.
\newblock \bibinfo{title}{Improving the performance of spiking neural networks on event-based datasets with knowledge transfer}.
\newblock \bibinfo{note}{ArXiv:2303.13077 [cs]}.
\bibitem[{Hidalgo-Carrió et~al.(2020)Hidalgo-Carrió, Gehrig and Scaramuzza}]{Hidalgo-Carrio2020Learning}
\bibinfo{author}{Hidalgo-Carrió, J.}, \bibinfo{author}{Gehrig, D.}, \bibinfo{author}{Scaramuzza, D.}, \bibinfo{year}{2020}.
\newblock \bibinfo{title}{Learning monocular dense depth from events}.
\newblock \bibinfo{note}{ArXiv:2010.08350 [cs]}.
\bibitem[{Hodgkin and Huxley(1952)}]{Hodgkin1952quantitative}
\bibinfo{author}{Hodgkin, A.L.}, \bibinfo{author}{Huxley, A.F.}, \bibinfo{year}{1952}.
\newblock \bibinfo{title}{A quantitative description of membrane current and its application to conduction and excitation in nerve}.
\newblock \bibinfo{note}{Publisher: Wiley-Blackwell}.
\bibitem[{Hsieh and Tang(2012)}]{hsieh2012vlsi}
\bibinfo{author}{Hsieh, H.Y.}, \bibinfo{author}{Tang, K.T.}, \bibinfo{year}{2012}.
\newblock \bibinfo{title}{Vlsi implementation of a bio-inspired olfactory spiking neural network}.
\newblock \bibinfo{journal}{IEEE transactions on neural networks and learning systems} \bibinfo{volume}{23}, \bibinfo{pages}{1065--1073}.
\bibitem[{Hu et~al.(2023)Hu, Tang and Pan}]{Hu2023Spiking}
\bibinfo{author}{Hu, Y.}, \bibinfo{author}{Tang, H.}, \bibinfo{author}{Pan, G.}, \bibinfo{year}{2023}.
\newblock \bibinfo{title}{Spiking deep residual networks}.
\newblock \DOIprefix\doi{10.1109/TNNLS.2021.3119238}. \bibinfo{note}{event-title: IEEE Transactions on Neural Networks and Learning Systems}.
\bibitem[{Huang et~al.(2022)Huang, Halwani, Muthusamy, Ayyad, Swart, Seneviratne, Gan and Zweiri}]{Huang2022Real-time}
\bibinfo{author}{Huang, X.}, \bibinfo{author}{Halwani, M.}, \bibinfo{author}{Muthusamy, R.}, \bibinfo{author}{Ayyad, A.}, \bibinfo{author}{Swart, D.}, \bibinfo{author}{Seneviratne, L.}, \bibinfo{author}{Gan, D.}, \bibinfo{author}{Zweiri, Y.}, \bibinfo{year}{2022}.
\newblock \bibinfo{title}{Real-time grasping strategies using event camera}.
\newblock \bibinfo{note}{Publisher: Springer}.
\bibitem[{Izhikevich(2003)}]{Izhikevich2003Simple}
\bibinfo{author}{Izhikevich, E.M.}, \bibinfo{year}{2003}.
\newblock \bibinfo{title}{Simple model of spiking neurons}.
\newblock \bibinfo{note}{Publisher: IEEE}.
\bibitem[{Johnson et~al.(2016)Johnson, Alahi and Fei-Fei}]{Johnson2016Perceptual}
\bibinfo{author}{Johnson, J.}, \bibinfo{author}{Alahi, A.}, \bibinfo{author}{Fei-Fei, L.}, \bibinfo{year}{2016}.
\newblock \bibinfo{title}{Perceptual losses for real-time style transfer and super-resolution}, in: \bibinfo{booktitle}{Computer Vision--ECCV 2016: 14th European Conference, Amsterdam, The Netherlands, October 11-14, 2016, Proceedings, Part II 14}, \bibinfo{organization}{Springer}. pp. \bibinfo{pages}{694--711}.
\bibitem[{Khan et~al.(2020)Khan, Salahuddin and Javidnia}]{Khan2020Deep}
\bibinfo{author}{Khan, F.}, \bibinfo{author}{Salahuddin, S.}, \bibinfo{author}{Javidnia, H.}, \bibinfo{year}{2020}.
\newblock \bibinfo{title}{Deep learning-based monocular depth estimation methods—a state-of-the-art review}.
\newblock \DOIprefix\doi{10.3390/s20082272}. \bibinfo{note}{number: 8 publisher: Multidisciplinary Digital Publishing Institute}.
\bibitem[{Kushawaha et~al.(2020)Kushawaha, Kumar, Banerjee and Velmurugan}]{Kushawaha2020Distilling}
\bibinfo{author}{Kushawaha, R.K.}, \bibinfo{author}{Kumar, S.}, \bibinfo{author}{Banerjee, B.}, \bibinfo{author}{Velmurugan, R.}, \bibinfo{year}{2020}.
\newblock \bibinfo{title}{Distilling spikes: Knowledge distillation in spiking neural networks}.
\newblock \DOIprefix\doi{10.48550/arXiv.2005.00288}. \bibinfo{note}{arXiv:2005.00288 [cs]}.
\bibitem[{Laga et~al.(2022)Laga, Jospin, Boussaid and Bennamoun}]{Laga2022Survey}
\bibinfo{author}{Laga, H.}, \bibinfo{author}{Jospin, L.V.}, \bibinfo{author}{Boussaid, F.}, \bibinfo{author}{Bennamoun, M.}, \bibinfo{year}{2022}.
\newblock \bibinfo{title}{A survey on deep learning techniques for stereo-based depth estimation}.
\newblock \DOIprefix\doi{10.1109/TPAMI.2020.3032602}. \bibinfo{note}{event-title: IEEE Transactions on Pattern Analysis and Machine Intelligence}.
\bibitem[{Laina et~al.(2016)Laina, Rupprecht, Belagiannis, Tombari and Navab}]{Laina2016Deeper}
\bibinfo{author}{Laina, I.}, \bibinfo{author}{Rupprecht, C.}, \bibinfo{author}{Belagiannis, V.}, \bibinfo{author}{Tombari, F.}, \bibinfo{author}{Navab, N.}, \bibinfo{year}{2016}.
\newblock \bibinfo{title}{Deeper depth prediction with fully convolutional residual networks}.
\newblock \bibinfo{note}{ArXiv:1606.00373 [cs] version: 2}.
\bibitem[{Lee et~al.(2020)Lee, Kosta, Zhu, Chaney, Daniilidis and Roy}]{Lee2020Spike-FlowNet}
\bibinfo{author}{Lee, C.}, \bibinfo{author}{Kosta, A.K.}, \bibinfo{author}{Zhu, A.Z.}, \bibinfo{author}{Chaney, K.}, \bibinfo{author}{Daniilidis, K.}, \bibinfo{author}{Roy, K.}, \bibinfo{year}{2020}.
\newblock \bibinfo{title}{Spike-flownet: Event-based optical flow estimation with energy-efficient hybrid neural networks}.
\newblock \DOIprefix\doi{10.1007/978-3-030-58526-6_22}.
\bibitem[{Lemaire et~al.(2022)Lemaire, Cordone, Castagnetti, Novac, Courtois and Miramond}]{Lemaire2022analytical}
\bibinfo{author}{Lemaire, E.}, \bibinfo{author}{Cordone, L.}, \bibinfo{author}{Castagnetti, A.}, \bibinfo{author}{Novac, P.E.}, \bibinfo{author}{Courtois, J.}, \bibinfo{author}{Miramond, B.}, \bibinfo{year}{2022}.
\newblock \bibinfo{title}{An analytical estimation of spiking neural networks energy efficiency}.
\bibitem[{Li et~al.(2020)Li, Zhu, Liu, Cao, Li, Jia and Qiu}]{Li2020Deep}
\bibinfo{author}{Li, Q.}, \bibinfo{author}{Zhu, J.}, \bibinfo{author}{Liu, J.}, \bibinfo{author}{Cao, R.}, \bibinfo{author}{Li, Q.}, \bibinfo{author}{Jia, S.}, \bibinfo{author}{Qiu, G.}, \bibinfo{year}{2020}.
\newblock \bibinfo{title}{Deep learning based monocular depth prediction: Datasets, methods and applications}.
\newblock \DOIprefix\doi{10.48550/arXiv.2011.04123}. \bibinfo{note}{arXiv:2011.04123 [cs]}.
\bibitem[{Liu et~al.(2022a)Liu, Zhang, Li, Lu, Huang and Zhang}]{Liu2022Unsupervised}
\bibinfo{author}{Liu, J.}, \bibinfo{author}{Zhang, Q.}, \bibinfo{author}{Li, J.}, \bibinfo{author}{Lu, M.}, \bibinfo{author}{Huang, T.}, \bibinfo{author}{Zhang, S.}, \bibinfo{year}{2022}a.
\newblock \bibinfo{title}{Unsupervised spike depth estimation via cross-modality cross-domain knowledge transfer}.
\newblock \DOIprefix\doi{10.48550/arXiv.2208.12527}. \bibinfo{note}{arXiv:2208.12527 [cs]}.
\bibitem[{Liu et~al.(2022b)Liu, Li, Fan and Tian}]{Liu2022Event-based}
\bibinfo{author}{Liu, X.}, \bibinfo{author}{Li, J.}, \bibinfo{author}{Fan, X.}, \bibinfo{author}{Tian, Y.}, \bibinfo{year}{2022}b.
\newblock \bibinfo{title}{Event-based monocular dense depth estimation with recurrent transformers}.
\newblock \bibinfo{note}{ArXiv:2212.02791 [cs]}.
\bibitem[{Liu et~al.(2021)Liu, Lin, Cao, Hu, Wei, Zhang, Lin and Guo}]{Liu2021Swin}
\bibinfo{author}{Liu, Z.}, \bibinfo{author}{Lin, Y.}, \bibinfo{author}{Cao, Y.}, \bibinfo{author}{Hu, H.}, \bibinfo{author}{Wei, Y.}, \bibinfo{author}{Zhang, Z.}, \bibinfo{author}{Lin, S.}, \bibinfo{author}{Guo, B.}, \bibinfo{year}{2021}.
\newblock \bibinfo{title}{Swin transformer: Hierarchical vision transformer using shifted windows}.
\newblock \bibinfo{note}{ArXiv: 2103.14030}.
\bibitem[{L{\"o}hr et~al.(2020)L{\"o}hr, Jarvers and Neumann}]{Lohr2020complex}
\bibinfo{author}{L{\"o}hr, M.P.}, \bibinfo{author}{Jarvers, C.}, \bibinfo{author}{Neumann, H.}, \bibinfo{year}{2020}.
\newblock \bibinfo{title}{Complex neuron dynamics on the ibm truenorth neurosynaptic system}.
\bibitem[{Mayr et~al.(2019)Mayr, Hoeppner and Furber}]{mayr2019spinnaker}
\bibinfo{author}{Mayr, C.}, \bibinfo{author}{Hoeppner, S.}, \bibinfo{author}{Furber, S.}, \bibinfo{year}{2019}.
\newblock \bibinfo{title}{Spinnaker 2: A 10 million core processor system for brain simulation and machine learning-keynote presentation}, in: \bibinfo{booktitle}{Communicating Process Architectures 2017 \& 2018}. \bibinfo{publisher}{IOS Press}, pp. \bibinfo{pages}{277--280}.
\bibitem[{Ming et~al.(2021)Ming, Meng, Fan and Yu}]{Ming2021Deep}
\bibinfo{author}{Ming, Y.}, \bibinfo{author}{Meng, X.}, \bibinfo{author}{Fan, C.}, \bibinfo{author}{Yu, H.}, \bibinfo{year}{2021}.
\newblock \bibinfo{title}{Deep learning for monocular depth estimation: A review}.
\newblock \DOIprefix\doi{10.1016/j.neucom.2020.12.089}.
\bibitem[{Nam et~al.(2022)Nam, Mostafavi, Yoon and Choi}]{Nam2022Stereo}
\bibinfo{author}{Nam, Y.}, \bibinfo{author}{Mostafavi, M.}, \bibinfo{author}{Yoon, K.J.}, \bibinfo{author}{Choi, J.}, \bibinfo{year}{2022}.
\newblock \bibinfo{title}{Stereo depth from events cameras: Concentrate and focus on the future}, \bibinfo{publisher}{IEEE}, \bibinfo{address}{New Orleans, LA, USA}. pp. \bibinfo{pages}{6104--6113}.
\newblock \DOIprefix\doi{10.1109/CVPR52688.2022.00602}. \bibinfo{note}{[Online; accessed 2023-08-18]}.
\bibitem[{Oquab et~al.(2023)Oquab, Darcet, Moutakanni, Vo, Szafraniec, Khalidov, Fernandez, Haziza, Massa, El-Nouby, Assran, Ballas, Galuba, Howes, Huang, Li, Misra, Rabbat, Sharma, Synnaeve, Xu, Jegou, Mairal, Labatut, Joulin and Bojanowski}]{Oquab2023DINOv2:}
\bibinfo{author}{Oquab, M.}, \bibinfo{author}{Darcet, T.}, \bibinfo{author}{Moutakanni, T.}, \bibinfo{author}{Vo, H.}, \bibinfo{author}{Szafraniec, M.}, \bibinfo{author}{Khalidov, V.}, \bibinfo{author}{Fernandez, P.}, \bibinfo{author}{Haziza, D.}, \bibinfo{author}{Massa, F.}, \bibinfo{author}{El-Nouby, A.}, \bibinfo{author}{Assran, M.}, \bibinfo{author}{Ballas, N.}, \bibinfo{author}{Galuba, W.}, \bibinfo{author}{Howes, R.}, \bibinfo{author}{Huang, P.Y.}, \bibinfo{author}{Li, S.W.}, \bibinfo{author}{Misra, I.}, \bibinfo{author}{Rabbat, M.}, \bibinfo{author}{Sharma, V.}, \bibinfo{author}{Synnaeve, G.}, \bibinfo{author}{Xu, H.}, \bibinfo{author}{Jegou, H.}, \bibinfo{author}{Mairal, J.}, \bibinfo{author}{Labatut, P.}, \bibinfo{author}{Joulin, A.}, \bibinfo{author}{Bojanowski, P.}, \bibinfo{year}{2023}.
\newblock \bibinfo{title}{Dinov2: Learning robust visual features without supervision}.
\newblock \DOIprefix\doi{10.48550/arXiv.2304.07193}. \bibinfo{note}{arXiv:2304.07193 [cs]}.
\bibitem[{Pehle et~al.(2022)Pehle, Billaudelle, Cramer, Kaiser, Schreiber, Stradmann, Weis, Leibfried, M{\"u}ller and Schemmel}]{Pehle2022brainscales}
\bibinfo{author}{Pehle, C.}, \bibinfo{author}{Billaudelle, S.}, \bibinfo{author}{Cramer, B.}, \bibinfo{author}{Kaiser, J.}, \bibinfo{author}{Schreiber, K.}, \bibinfo{author}{Stradmann, Y.}, \bibinfo{author}{Weis, J.}, \bibinfo{author}{Leibfried, A.}, \bibinfo{author}{M{\"u}ller, E.}, \bibinfo{author}{Schemmel, J.}, \bibinfo{year}{2022}.
\newblock \bibinfo{title}{The brainscales-2 accelerated neuromorphic system with hybrid plasticity}.
\bibitem[{Posch et~al.(2011)Posch, Matolin and Wohlgenannt}]{Posch2011QVGA}
\bibinfo{author}{Posch, C.}, \bibinfo{author}{Matolin, D.}, \bibinfo{author}{Wohlgenannt, R.}, \bibinfo{year}{2011}.
\newblock \bibinfo{title}{A qvga 143 db dynamic range frame-free pwm image sensor with lossless pixel-level video compression and time-domain cds}.
\newblock \DOIprefix\doi{10.1109/JSSC.2010.2085952}. \bibinfo{note}{event-title: IEEE Journal of Solid-State Circuits}.
\bibitem[{Qiu et~al.(2022)Qiu, Ning, Yuan, Fang, Chen, Li, Sun, Ma and Tian}]{Qiu2022Self-Architectural}
\bibinfo{author}{Qiu, H.}, \bibinfo{author}{Ning, M.}, \bibinfo{author}{Yuan, L.}, \bibinfo{author}{Fang, W.}, \bibinfo{author}{Chen, Y.}, \bibinfo{author}{Li, C.}, \bibinfo{author}{Sun, T.}, \bibinfo{author}{Ma, Z.}, \bibinfo{author}{Tian, Y.}, \bibinfo{year}{2022}.
\newblock \bibinfo{title}{Self-architectural knowledge distillation for spiking neural networks}.
\newblock \bibinfo{note}{[Online; accessed 2023-08-22]}.
\bibitem[{Rafi(2021)}]{Rafi2021brief}
\bibinfo{author}{Rafi, T.H.}, \bibinfo{year}{2021}.
\newblock \bibinfo{title}{A brief review on spiking neural network-a biological inspiration}.
\newblock \bibinfo{note}{Publisher: Preprints}.
\bibitem[{Ranftl et~al.(2021)Ranftl, Bochkovskiy and Koltun}]{Ranftl2021Vision}
\bibinfo{author}{Ranftl, R.}, \bibinfo{author}{Bochkovskiy, A.}, \bibinfo{author}{Koltun, V.}, \bibinfo{year}{2021}.
\newblock \bibinfo{title}{Vision transformers for dense prediction}.
\newblock \bibinfo{note}{ArXiv:2103.13413 [cs]}.
\bibitem[{Ranftl et~al.(2020)Ranftl, Lasinger, Hafner, Schindler and Koltun}]{Ranftl2020Towards}
\bibinfo{author}{Ranftl, R.}, \bibinfo{author}{Lasinger, K.}, \bibinfo{author}{Hafner, D.}, \bibinfo{author}{Schindler, K.}, \bibinfo{author}{Koltun, V.}, \bibinfo{year}{2020}.
\newblock \bibinfo{title}{Towards robust monocular depth estimation: Mixing datasets for zero-shot cross-dataset transfer}.
\newblock \DOIprefix\doi{10.48550/arXiv.1907.01341}. \bibinfo{note}{arXiv:1907.01341 [cs]}.
\bibitem[{Ronneberger et~al.(2015)Ronneberger, Fischer and Brox}]{Ronneberger2015U-Net:}
\bibinfo{author}{Ronneberger, O.}, \bibinfo{author}{Fischer, P.}, \bibinfo{author}{Brox, T.}, \bibinfo{year}{2015}.
\newblock \bibinfo{title}{U-net: Convolutional networks for biomedical image segmentation}, pp. \bibinfo{pages}{234--241}.
\bibitem[{Strudel et~al.(2021)Strudel, Garcia, Laptev and Schmid}]{Strudel2021Segmenter:}
\bibinfo{author}{Strudel, R.}, \bibinfo{author}{Garcia, R.}, \bibinfo{author}{Laptev, I.}, \bibinfo{author}{Schmid, C.}, \bibinfo{year}{2021}.
\newblock \bibinfo{title}{Segmenter: Transformer for semantic segmentation}, pp. \bibinfo{pages}{7262--7272}.
\newblock \bibinfo{note}{[Online; accessed 2023-08-15]}.
\bibitem[{Sun et~al.(2021)Sun, Cao, Yang and Kitani}]{Sun2021Rethinking}
\bibinfo{author}{Sun, Z.}, \bibinfo{author}{Cao, S.}, \bibinfo{author}{Yang, Y.}, \bibinfo{author}{Kitani, K.M.}, \bibinfo{year}{2021}.
\newblock \bibinfo{title}{Rethinking transformer-based set prediction for object detection}, pp. \bibinfo{pages}{3611--3620}.
\newblock \bibinfo{note}{[Online; accessed 2023-08-15]}.
\bibitem[{Tavanaei et~al.(2019)Tavanaei, Ghodrati, Kheradpisheh, Masquelier and Maida}]{tavanaei2019deep}
\bibinfo{author}{Tavanaei, A.}, \bibinfo{author}{Ghodrati, M.}, \bibinfo{author}{Kheradpisheh, S.R.}, \bibinfo{author}{Masquelier, T.}, \bibinfo{author}{Maida, A.}, \bibinfo{year}{2019}.
\newblock \bibinfo{title}{Deep learning in spiking neural networks}.
\newblock \bibinfo{journal}{Neural networks} \bibinfo{volume}{111}, \bibinfo{pages}{47--63}.
\bibitem[{Tayarani-Najaran and Schmuker(2021)}]{Tayarani-Najaran2021Event-Based}
\bibinfo{author}{Tayarani-Najaran, M.H.}, \bibinfo{author}{Schmuker, M.}, \bibinfo{year}{2021}.
\newblock \bibinfo{title}{Event-based sensing and signal processing in the visual, auditory, and olfactory domain: a review}.
\newblock \bibinfo{journal}{Frontiers in Neural Circuits} \bibinfo{volume}{15}, \bibinfo{pages}{610446}.
\bibitem[{Wang et~al.(2021)Wang, Xie, Li, Fan, Song, Liang, Lu, Luo and Shao}]{Wang2021Pyramid}
\bibinfo{author}{Wang, W.}, \bibinfo{author}{Xie, E.}, \bibinfo{author}{Li, X.}, \bibinfo{author}{Fan, D.P.}, \bibinfo{author}{Song, K.}, \bibinfo{author}{Liang, D.}, \bibinfo{author}{Lu, T.}, \bibinfo{author}{Luo, P.}, \bibinfo{author}{Shao, L.}, \bibinfo{year}{2021}.
\newblock \bibinfo{title}{Pyramid vision transformer: A versatile backbone for dense prediction without convolutions}.
\newblock \bibinfo{note}{[Online; accessed 2021-05-06]}.
\bibitem[{Yang et~al.(2022)Yang, An, Dixit, Koo and Park}]{Yang2022Depth}
\bibinfo{author}{Yang, J.}, \bibinfo{author}{An, L.}, \bibinfo{author}{Dixit, A.}, \bibinfo{author}{Koo, J.}, \bibinfo{author}{Park, S.I.}, \bibinfo{year}{2022}.
\newblock \bibinfo{title}{Depth estimation with simplified transformer}.
\newblock \DOIprefix\doi{10.48550/arXiv.2204.13791}. \bibinfo{note}{arXiv:2204.13791 [cs]}.
\bibitem[{Yao et~al.(2023)Yao, Hu, Zhou, Yuan, Tian, Xu and Li}]{Yao2023Spike-driven}
\bibinfo{author}{Yao, M.}, \bibinfo{author}{Hu, J.}, \bibinfo{author}{Zhou, Z.}, \bibinfo{author}{Yuan, L.}, \bibinfo{author}{Tian, Y.}, \bibinfo{author}{Xu, B.}, \bibinfo{author}{Li, G.}, \bibinfo{year}{2023}.
\newblock \bibinfo{title}{Spike-driven transformer}.
\newblock \DOIprefix\doi{10.48550/arXiv.2307.01694}. \bibinfo{note}{arXiv:2307.01694 [cs]}.
\bibitem[{Yin et~al.(2020)Yin, Corradi and Bohté}]{Yin2020Effective}
\bibinfo{author}{Yin, B.}, \bibinfo{author}{Corradi, F.}, \bibinfo{author}{Bohté, S.M.}, \bibinfo{year}{2020}.
\newblock \bibinfo{title}{Effective and efficient computation with multiple-timescale spiking recurrent neural networks}, \bibinfo{publisher}{Association for Computing Machinery}, \bibinfo{address}{New York, NY, USA}. p. \bibinfo{pages}{1–8}.
\newblock \DOIprefix\doi{10.1145/3407197.3407225}. \bibinfo{note}{[Online; accessed 2023-08-15]}.
\bibitem[{Zhang et~al.(2022)Zhang, Tang, Yu, Lu and Huang}]{Zhang2022Spike}
\bibinfo{author}{Zhang, J.}, \bibinfo{author}{Tang, L.}, \bibinfo{author}{Yu, Z.}, \bibinfo{author}{Lu, J.}, \bibinfo{author}{Huang, T.}, \bibinfo{year}{2022}.
\newblock \bibinfo{title}{Spike transformer: Monocular depth estimation for spiking camera}, in: \bibinfo{booktitle}{European Conference on Computer Vision}, \bibinfo{organization}{Springer}. pp. \bibinfo{pages}{34--52}.
\bibitem[{Zhao et~al.(2020)Zhao, Sun, Zhang, Tang and Qian}]{Zhao2020Monocular}
\bibinfo{author}{Zhao, C.}, \bibinfo{author}{Sun, Q.}, \bibinfo{author}{Zhang, C.}, \bibinfo{author}{Tang, Y.}, \bibinfo{author}{Qian, F.}, \bibinfo{year}{2020}.
\newblock \bibinfo{title}{Monocular depth estimation based on deep learning: An overview}.
\newblock \DOIprefix\doi{10.1007/s11431-020-1582-8}.
\bibitem[{Zhao et~al.(2022)Zhao, Zhang, Poggi, Tosi, Guo, Zhu, Huang, Tang and Mattoccia}]{Zhao2022MonoViT:}
\bibinfo{author}{Zhao, C.}, \bibinfo{author}{Zhang, Y.}, \bibinfo{author}{Poggi, M.}, \bibinfo{author}{Tosi, F.}, \bibinfo{author}{Guo, X.}, \bibinfo{author}{Zhu, Z.}, \bibinfo{author}{Huang, G.}, \bibinfo{author}{Tang, Y.}, \bibinfo{author}{Mattoccia, S.}, \bibinfo{year}{2022}.
\newblock \bibinfo{title}{Monovit: Self-supervised monocular depth estimation with a vision transformer}, pp. \bibinfo{pages}{668--678}.
\newblock \DOIprefix\doi{10.1109/3DV57658.2022.00077}. \bibinfo{note}{iSSN: 2475-7888}.
\bibitem[{Zheng et~al.(2020)Zheng, Wu, Deng, Hu and Li}]{Zheng2020Going}
\bibinfo{author}{Zheng, H.}, \bibinfo{author}{Wu, Y.}, \bibinfo{author}{Deng, L.}, \bibinfo{author}{Hu, Y.}, \bibinfo{author}{Li, G.}, \bibinfo{year}{2020}.
\newblock \bibinfo{title}{Going deeper with directly-trained larger spiking neural networks}.
\newblock \DOIprefix\doi{10.48550/arXiv.2011.05280}. \bibinfo{note}{arXiv:2011.05280 [cs]}.
\bibitem[{Zhou et~al.(2023)Zhou, Yu, Zhou, Ma, Zhang, Zhou and Tian}]{Zhou2023Spikingformer:}
\bibinfo{author}{Zhou, C.}, \bibinfo{author}{Yu, L.}, \bibinfo{author}{Zhou, Z.}, \bibinfo{author}{Ma, Z.}, \bibinfo{author}{Zhang, H.}, \bibinfo{author}{Zhou, H.}, \bibinfo{author}{Tian, Y.}, \bibinfo{year}{2023}.
\newblock \bibinfo{title}{Spikingformer: Spike-driven residual learning for transformer-based spiking neural network}.
\newblock \bibinfo{note}{ArXiv:2304.11954 [cs]}.
\bibitem[{Zhou et~al.(2022)Zhou, Zhu, He, Wang, Yan, Tian and Yuan}]{Zhou2022Spikformer:}
\bibinfo{author}{Zhou, Z.}, \bibinfo{author}{Zhu, Y.}, \bibinfo{author}{He, C.}, \bibinfo{author}{Wang, Y.}, \bibinfo{author}{Yan, S.}, \bibinfo{author}{Tian, Y.}, \bibinfo{author}{Yuan, L.}, \bibinfo{year}{2022}.
\newblock \bibinfo{title}{Spikformer: When spiking neural network meets transformer}.
\newblock \DOIprefix\doi{10.48550/arXiv.2209.15425}. \bibinfo{note}{arXiv:2209.15425 [cs]}.

\end{thebibliography}

\end{document}